\documentclass[preprint,10pt]{elsarticle}
\usepackage{subcaption}
\usepackage[table]{xcolor}

\usepackage{graphicx}%
\usepackage{multirow}%
\usepackage{amsmath,amssymb,amsfonts}%
\usepackage{amsthm}%
\usepackage{mathrsfs}%
\usepackage[title]{appendix}%
\usepackage{xcolor}%
\usepackage{textcomp}%
\usepackage{manyfoot}%
\usepackage{booktabs}%
\usepackage{algorithm}%
\usepackage{algorithmicx}%
\usepackage{algpseudocode}%
\usepackage{listings}%
\usepackage{rotating}
\usepackage{booktabs}
\usepackage{tabularx}

\journal{Reliability Engineering \& System Safety}

\definecolor{RED}{RGB}{191, 25, 50}
\definecolor{YELLOW}{RGB}{245, 223, 77}
\definecolor{BLUE}{RGB}{15, 76, 129}

\makeatletter
\def\ps@pprintTitle{%
 \let\@oddhead\@empty
 \let\@evenhead\@empty
 \let\@oddfoot\@empty
 \let\@evenfoot\@empty
}
\makeatother

\begin{document}

\begin{frontmatter}

\title{Robust and Safe Traffic Sign Recognition using N-version with Weighted Voting}

\author[inst1]{Linyun Gao}
\ead{gao.linyun@sd.cs.tsukuba.ac.jp}

\author[inst1]{Qiang Wen\corref{cor1}}
\ead{wen.qiang@sd.cs.tsukuba.ac.jp}

\author[inst1]{Fumio Machida}
\ead{machida@cs.tsukuba.ac.jp}

\cortext[cor1]{Corresponding author}

\affiliation[inst1]{organization={Department of Computer Science, University of Tsukuba},
            addressline={1-1-1 Tennodai}, 
            city={Tsukuba},
            postcode={305-8573}, 
            state={Ibaraki},
            country={Japan}}




\begin{abstract}
Autonomous driving is rapidly advancing as a key application of machine learning, yet ensuring the safety of these systems remains a critical challenge. Traffic sign recognition, an essential component of autonomous vehicles, is particularly vulnerable to adversarial attacks that can compromise driving safety. In this paper, we propose an N-version machine learning (NVML) framework that integrates a safety-aware weighted soft voting mechanism. Our approach utilizes Failure Mode and Effects Analysis (FMEA) to assess potential safety risks and assign dynamic, safety-aware weights to the ensemble outputs.  We evaluate the robustness of three-version NVML systems employing various voting mechanisms against adversarial samples generated using the Fast Gradient Sign Method (FGSM) and Projected Gradient Descent (PGD) attacks. Experimental results demonstrate that our NVML approach significantly enhances the robustness and safety of traffic sign recognition systems under adversarial conditions.

\end{abstract}


\end{frontmatter}

\newpage

\section{Introduction}
\label{sec:intro}

In recent years, machine learning (ML) has become a fundamental technology in safety-critical domains, including autonomous driving. In particular, computer vision (CV) systems powered by ML models play a crucial role in enabling autonomous vehicles to perceive and interpret their surroundings \cite{dilek2023computer}. However, many state-of-the-art CV models rely on complex, unexplainable black-box ML models such as deep neural networks (DNNs). The uncertain outputs raise concerns about the reliability of such systems, especially in high-risk environments where incorrect predictions can lead to severe consequences. Therefore, developing robust methods to enhance the reliability of ML systems remains an urgent research challenge.

The N-version Machine Learning (NVML) system leverages the traditional N-version programming technique by incorporating diverse ML models within the same system to enhance output reliability \citep{machida2019n}. Prior research has shown that NVML can improve system reliability in tasks such as image classification \citep{xu2019nv} and steering control \citep{wu2018model}. Since different models in an NVML system may produce varying predictions, an effective decision-making mechanism is required to ensure reliable outputs. For example, in a three-version ML-based image classification system, three independent models generate independent predictions, and the final result is determined through majority voting \citep{wen2022reliability}.

However, majority voting has inherent limitations. A major drawback is its inability to resolve cases where all models produce distinct predictions, resulting in a tie with no clear consensus. For example, if three models predict classes A, B, and C, the system is unable to determine a definitive outcome. A simple majority vote may not always yield the most accurate result. In such scenarios, prioritizing the prediction from the model with the highest historical accuracy could be a more effective strategy, as it is more likely to provide a reliable classification for difficult cases. 

In this work, we explore more refined voting mechanisms for NVML systems in safety-critical applications. We investigate multiple decision strategies in the NVML image classification systems, including majority voting, weighted voting, soft voting, and weighted soft voting. In particular, we emphasize the weighted soft voting approach, which leverages the confidence levels of model predictions to assign appropriate weights to each model. By incorporating confidence scores and reliability of different models, this method has the potential to enable the system to make more informed decisions, ultimately enhancing system reliability in safety-critical applications.
To determine safety-aware weights in safety-critical applications, we introduce a metric to evaluate the safety of ML-based traffic sign recognition systems for autonomous vehicles. We utilize the Failure Mode and Effects Analysis (FMEA) method to systematically examine potential failures arising from recognition errors during autonomous driving. By identifying and assessing these failure modes, we assign risk scores based on severity and likelihood. These scores guide the weighting process, ensuring that models associated with lower risks have a greater influence on the system’s final output.

We evaluate the effectiveness of various voting mechanisms, including weighted soft voting, through experiments on a traffic sign recognition system comprising three distinct ML models: AlexNet \citep{krizhevsky2012imagenet}, VGG16 \citep{simonyan2014very}, and EfficientNetB0 \citep{tan2019efficientnet}. System performance is assessed using metrics such as accuracy, safety score, and the number of high-severity misclassifications. These metrics provide a comparative analysis of different voting strategies.
Our results demonstrate that the proposed weighted soft voting method not only enhances accuracy but also significantly improves the safety and reliability of the NVML system by reducing severe misclassification instances. Additionally, we introduce an LLM-based weight assignment approach, where ChatGPT generates model weights based on a severity matrix and misclassification probabilities. However, preliminary evaluations indicate that this method produces suboptimal solutions, highlighting the need for more specialized LLMs and improved prompt engineering.

The existing study \cite{gao2024safety} has demonstrated the advantage of safety-aware weighted voting for N-version traffic sign recognition systems; we extend this work by analyzing the effectiveness of NVML systems in adversarial environments. The NVML system with the proposed voting mechanism can improve the reliability even under adversarial attack conditions. Adversarial attacks pose a significant challenge to the reliability of ML systems, particularly in safety-critical applications such as autonomous vehicles. These attacks exploit vulnerabilities in ML models by introducing imperceptible perturbations, leading to potentially severe consequences. While traditional single-model systems often lack sufficient robustness against such attacks, NVML systems leverage model redundancy and diversity to enhance robustness. We evaluate the robustness of both single-model ML systems and a proposed three-version NVML system comprising AlexNet, VGG16, and EfficientNetB0, using adversarial samples generated with the Fast Gradient Sign Method (FGSM) \citep{goodfellow2014explaining} and Projected Gradient Descent (PGD) \citep{madry2017towards} attacks. The results reveal that while single-model systems experience significant performance degradation, the NVML system demonstrates superior robustness, particularly when employing advanced voting mechanisms such as safety-aware weighted soft voting. These findings underscore the critical role of NVML systems in ensuring safety and reliability in real-world, even under adversarial environments.

The contributions of this paper can be summarized as follows:

\begin{itemize}
    \item We introduce a safety-aware weighted soft voting mechanism for NVML systems.
    \item We develop a safety metric based on the FMEA method to evaluate and improve the safety of ML-based traffic sign recognition systems.
    \item We demonstrate the effectiveness of the proposed safety-aware weighted voting mechanism through the experiments using a real traffic sign dataset. 
    \item We examine the potential LLM-based approach for weight assignment.
    \item Our extensive robustness experiments show that the proposed NVML system, especially when using weighted soft voting, significantly enhances reliability and resilience under adversarial attacks, outperforming single-model systems in maintaining high accuracy and safety.
\end{itemize}

The rest of the paper is organized as follows: Section \ref{sec:related_work} reviews related work. Section \ref{sec:background} provides background information on traffic sign recognition and the FMEA approach. Section \ref{sec:voting_mechanism} describes the proposed voting mechanisms. Section \ref{sec:safety_metric} presents the proposed safety metric. Section \ref{sec:experiment} presents the experimental setup and results, and Section \ref{sec:conclusion} concludes the paper.

\section{Related work}\label{sec:related_work}

Recent advancements in multi-version ML techniques have aimed at enhancing the reliability of ML systems. 
Xu et al. explored the potential of developing fault-tolerant deep learning systems through model redundancy (i.e., NV-DNN \citep{xu2019nv}). Experimental results indicate that their approach can improve the fault tolerance of deep learning systems. Unlike NV-DNN, which processes only a single input at a time, N-version ML can incorporate multiple inputs, thereby leveraging input diversity to further improve reliability. 
Makino et al. \cite{makino2021queueing} developed queueing models to analyze the performance of multi-model multi-input ML systems, particularly focusing on the throughput of two-version architectures, where up to two ML models and two data sources are utilized. 
Wen and Machida \cite{wen2022reliability, wen2023characterizing, wen2025reliability} conducted numerical and empirical analysis on the impact of using diverse models and varied inputs to enhance the reliability of three-version ML systems. Hong et al. \cite{hong2020more} introduced a multimodal deep learning method that enhances the classification accuracy of remote-sensing imagery, outperforming the performance of both single-model or single-modality approaches.

Multiple voting strategies have been explored in multi-version ML approaches.
Singamsetty and Panchumarthy \cite{singamsetty2011novel} analyzed existing weighted average voting methods in safety-critical applications and proposed a history-based weighted voting algorithm featuring a soft dynamic threshold. 
Karimi et al. \cite{karimi2014novel} introduced a voting algorithm for real-time fault-tolerant control systems, especially for large N. Their approach addresses the limitations of median and weighted voting algorithms, significantly improving system reliability and availability.
Wu et al. \cite{wu2018model} developed a weighted N-version programming scheme to enhance the resilience of ML-based steering control algorithms.
The design of the scheme is based on the fusion of three redundant DNN model outputs. They proposed a weighted voting scheme based on the steering angle RMSE performance. 

Our study focuses on the impact of safety-aware weight assignment, considering misclassification severity in NVML traffic sign classification systems.

Existing studies have explored safety-aware strategies to improve system reliability in critical applications. 
Zhao et al. \cite{zhao2019towards} examined safety-aware computing system design for autonomous vehicles, introducing a safety score metric that extends beyond traditional performance evaluations. They also developed a perception latency model to estimate safety scores, demonstrating its application in hardware resource management for enhanced safety in CV systems used in autonomous driving. Rahman et al. \cite{rahman2024investigating} analyzed the impact of transient hardware faults on the misclassification of DNN models based on safety-critical metrics compared to intrinsic algorithmic inaccuracy. 
Additionally, Gao et al. \cite{gao2024safety} introduces the safety-aware weighted voting N-version traffic sign recognition system to improve the safety of ML-based systems. In this work, we further illustrate the effectiveness of NVML systems in adversarial environments.

Adversarial attacks have emerged as a significant concern in ML applications, where even imperceptible perturbations to input data can lead to incorrect predictions. Pioneering studies, such as those by Goodfellow et al. \cite{goodfellow2014explaining}, introduced FGSM to demonstrate the vulnerability of ML models to adversarial perturbations. 
Subsequent work expanded on this by developing more sophisticated attack strategies, including iterative methods like PGD \citep{madry2017towards} and black-box approaches \citep{papernot2017practical}. These attacks have highlighted critical weaknesses in CV models, such as image classifiers and object detectors, which are often deployed in safety-critical applications like autonomous driving and surveillance. While defenses like adversarial training \citep{madry2017towards} have shown promise, achieving robust and generalized protection against adversarial attacks remains an open challenge, driving ongoing research in understanding and mitigating vulnerabilities in ML and CV systems.

\section{Background}\label{sec:background}

\subsection{Traffic sign recognition}



For autonomous vehicles, the precise identification of traffic signs is a cornerstone of safe operation. These vehicles rely on systems that can rapidly detect and interpret regulatory, warning, and informational signs to follow road rules and maintain safety. Typically, such systems process live video feeds from onboard cameras using advanced machine learning methods.

\begin{figure}[h]
    \centering
    \begin{subfigure}[b]{0.4\textwidth}
        \centering
        \includegraphics[height=2cm]{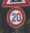}
        \caption{Speed Limit 20 sign frequently misclassified as Speed Limit 70 sign}
        \label{fig:sl20}
    \end{subfigure}
    \hspace{0.06\textwidth}
    \begin{subfigure}[b]{0.4\textwidth}
        \centering
        \includegraphics[height=2cm]{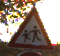}
        \caption{School Crossing sign frequently misclassified as Cycles Crossing sign}
        \label{fig:sc}
    \end{subfigure}
    \caption{Frequently Misclassified Traffic Signs}
    \label{fig:traffic_sign}
\end{figure}

Errors in this recognition process can lead to severe consequences. For example, if a system misreads a "Speed Limit 20" sign as a "Speed Limit 70" sign (see Fig. \ref{fig:sl20}), the vehicle might accelerate beyond safe limits. In contrast, confusing a "School Crossing" sign with a "Cycles Crossing" sign (see Fig. \ref{fig:sc}) is generally less dangerous, as both serve to alert drivers. These scenarios demonstrate the critical importance of developing traffic sign recognition systems that are both accurate and reliable. Our investigation, which includes experiments with models such as AlexNet, has uncovered both high-risk and relatively benign misinterpretations, underscoring the need for continual system enhancement.

\subsection{Failure modes and effects analysis}





FMEA offers a structured strategy to identify potential failure points in a system and evaluate their impacts \citep{khaiyum2015approach}. This method is widely adopted in industries like automotive manufacturing, engineering, and healthcare to preemptively address risks before they lead to problems.

Our research focuses on enhancing the safety of machine learning-based computer vision systems in autonomous vehicles. These systems must reliably recognize a variety of objects—ranging from other vehicles and pedestrians to traffic signals and signs—to ensure safe navigation. In particular, we concentrate on the challenges inherent in traffic sign recognition.

We utilize FMEA to rigorously assess the risks associated with misclassifications in traffic sign detection across different driving scenarios. For instance, misinterpreting a 60 km/h speed limit as 120 km/h could induce dangerous speeding. The FMEA framework allows us to systematically rank these risks and target the most critical vulnerabilities.

The FMEA procedure in our work involves three primary steps:

    
    






\begin{enumerate}[1.]
    \item \textbf{FMEA Worksheet Development}
    
    We construct a detailed worksheet that enumerates each function of the system, the possible failure modes, and their anticipated effects.
    
    \item \textbf{Evaluation of Failure Effects}
    
    Each failure mode is analyzed by considering:
    \begin{itemize}
        \item \textbf{Probability (P)}: The likelihood of the failure occurring.
        \item \textbf{Severity (S)}: The potential impact if the failure occurs.
        \item \textbf{Detection (D)}: The chance of detecting the failure before it causes harm.
    \end{itemize}
    
    Given that our traffic sign recognition system operates in real time with images from vehicle cameras, we exclude the detection parameter from our evaluation.
    
    \item \textbf{Risk Prioritization}
    
    We calculate the risk associated with each failure using the equation $R = P \times S$. This metric helps in identifying which issues demand immediate attention, thereby guiding targeted improvements to enhance system safety.
\end{enumerate}

\section{Voting mechanism}\label{sec:voting_mechanism}


The NVML framework aggregates predictions from several machine learning models. A robust voting scheme is essential to fuse these individual predictions into a trustworthy final decision. In our exploration of traffic sign recognition, we evaluated various voting strategies. Depending on the type of outputs generated by the models, these strategies are classified into hard voting and soft voting.

\subsection{Hard voting}


Hard voting relies on the discrete class outputs provided by each model. Every model produces a single class label, and these labels are combined to determine the overall result. Assume that the NVML framework is composed of $T$ independent models, denoted as $\{h_1, h_2, \dots, h_T\}$, and that the task involves $l$ possible classes $\{c_1, c_2, \dots, c_l\}$. For any input $x$, the prediction from the $i$-th model, $h_i$, is defined as:

\begin{equation}
h_i(x) = \begin{bmatrix} h_{i1}(x) & h_{i2}(x) & \dots & h_{il}(x) \end{bmatrix}^T
\end{equation}


where $h_{ij}(x)$ is given by

\begin{equation}
    h_{ij}(x) = 
\begin{cases} 
1 & \text{if model } i \text{ predicts class } j, \\
0 & \text{otherwise.}
\end{cases}
\end{equation}


Hard voting can be implemented using either majority voting or weighted voting.

\subsubsection{Majority voting}


In majority voting, the class that garners the highest number of votes is chosen as the final outcome. If multiple classes receive an equal highest vote count, no decision is rendered.
The final decision $H_m(x)$ is formulated as:

\begin{equation}
H_m(x) = 
\begin{cases} 
c_k, & \text{if } \exists \, k \in \{1, 2, \dots, l\}, \, \sum_{i=1}^T h_{ik}(x) > \frac{T}{2}, \\
\text{no output}, & \text{otherwise.}
\end{cases}
\end{equation}


Although majority voting is intuitive and simple to execute, it fails to differentiate between models of varying accuracy, potentially allowing less reliable predictions to influence the result.  

\subsubsection{Weighted voting}



Weighted voting assigns a specific importance weight to each model. The final prediction is determined by the class that accumulates the highest weighted sum of votes.

The system output $H_w(x)$ is defined as:

\begin{equation}
    H_w(x) = c_k, \quad k = \text{argmax}_{j \in \{1, 2, \dots, l\}} \sum_{i=1}^T w_i \cdot h_{ij}(x)
\end{equation}


subject to the constraints:

\begin{equation}
    \sum_{i=1}^T w_i = 1 \quad \text{and} \quad w_i > 0 \quad \forall i.
\end{equation}



Here, $w_i$ denotes the weight attributed to the $i$-th model, $h_i$. This method adjusts for differences in model reliability but selecting optimal weights can be intricate and might introduce biases if certain models are overemphasized.

\subsection{Soft voting}


Soft voting, in contrast, uses the probability outputs from each model. Every model returns a probability distribution over all classes, and these probabilities are aggregated to decide the final output. Assume that the NVML system includes $T$ independent models $\{g_1, g_2, \dots, g_T\}$, and that the classification task involves $l$ classes $\{c_1, c_2, \dots, c_l\}$. For an input $x$, the prediction from the $i$-th model, $g_i$, is expressed as:

\begin{equation}
    g_i(x) = \begin{bmatrix} g_{i1}(x) & g_{i2}(x) & \dots & g_{il}(x) \end{bmatrix}^T
\end{equation}


where $g_{ij}(x)$ represents the probability that model $i$ assigns to class $c_j$, with the constraints

\begin{equation}
    g_{ij}(x) \geq 0 \quad \text{and} \quad \sum_{j=1}^l g_{ij}(x) = 1
\end{equation}


Soft voting is divided into two approaches: simple soft voting and weighted soft voting.

\subsubsection{Simple soft voting}


Simple soft voting aggregates the probability vectors from all models, and the class with the highest total probability is selected as the final output $G_s(x)$ \citep{xu2019nv}. This is expressed mathematically as:


\begin{equation}
    G_s(x) = c_k, \quad k = \text{argmax}_{j \in \{1, 2, \dots, l\}}  \sum_{i=1}^T g_{ij}(x)
\end{equation}


This method leverages the entire probability distribution to capture the models' confidence. Nonetheless, the numerical probability scores may not always be well-calibrated, meaning high probabilities do not invariably indicate correctness, which could affect reliability.

\subsubsection{Weighted soft voting}


Weighted soft voting enhances soft voting by incorporating model-specific weights. The final prediction $G_w(x)$ is the class with the highest sum of weighted probabilities \citep{wu2018model}:


\begin{equation}
    G_w(x) = c_k, \quad k = \text{argmax}_{j \in \{1, 2, \dots, l\}} \sum_{i=1}^T w_i \cdot g_{ij}(x)
\end{equation}

subject to:

\begin{equation}
    \sum_{i=1}^T w_i = 1 \quad \text{and} \quad w_i > 0 \quad \forall i
\end{equation}



In this context, $w_i$ represents the normalized weight assigned to model $g_i$, indicating its significance or trustworthiness. While this approach fuses the benefits of both weighted and soft voting, the challenge of selecting suitable weights persists.

\subsection{Weight Assignment Challenges}


Taking into account the benefits and drawbacks of the various voting methods outlined in Table \ref{tab:voting_mechanism}, we have chosen weighted soft voting for our study. This strategy merges the advantages of assigning model-specific weights with the detailed confidence information provided by probability distributions. Nevertheless, determining optimal weights remains problematic due to the lack of a standardized procedure. This issue drives our investigation into innovative methods for assigning weights to individual ML models in order to bolster system safety. To this end, we propose a safety metric that not only assesses system robustness but also informs the weight allocation process within the NVML framework.

\begin{table}[h]\scriptsize
    \caption{Comparison of voting mechanisms}\label{tab:voting_mechanism}
    \begin{tabularx}{\textwidth}{@{}lp{42mm}p{44mm}}
        \toprule
        Voting Mechanism & Advantage & Disadvantage \\
        \midrule
        Majority Voting & Easy to comprehend and implement; relies solely on class labels, making it computationally efficient and simple. & Ignores differences in model accuracy, which can allow less reliable predictions to unduly sway the final decision. \\
        Weighted Voting & Adjusts for differences in model performance by assigning distinct weights, potentially improving overall accuracy. & Determining the optimal weights is challenging and may require extensive experimentation, increasing complexity. \\
        Simple Soft Voting & Incorporates complete probability distributions, thereby reflecting the confidence of predictions and handling uncertainty better. & The probability estimates might not accurately represent true confidence levels, potentially leading to erroneous decisions. \\
        Weighted Soft Voting & Combines the merits of weighting and soft voting by using both model-specific weights and probabilistic outputs, resulting in a more refined decision process. & Similar to weighted voting, it faces the challenge of choosing appropriate weights, which can be computationally demanding in complex systems. \\
        \bottomrule
    \end{tabularx}
\end{table}

\section{Safety metric}\label{sec:safety_metric}



A traffic sign recognition system is indispensable for autonomous vehicles, as its dependability directly influences overall vehicle safety. To bolster safety, we have developed a specialized safety metric for ML-based traffic sign recognition systems. This metric not only assesses the performance of individual ML models and the integrated NVML system but also underpins the weight assignment in our weighted voting strategies.

In our work, we utilize the publicly available GTSRB \cite{stallkamp2012man} and ArTS \cite{alghmgham2019autonomous} datasets and employ the FMEA methodology to uncover potential risks within traffic sign recognition systems. The insights obtained from this risk analysis enable us to evaluate the safety of the ML models and the overall system, which, in turn, informs the weight determination for the weighted voting mechanisms.

\subsection{Failure modes and effects analysis}

\subsubsection{FMEA Worksheet}


Table \ref{tab:fmea_worksheet} presents an FMEA worksheet that enumerates all conceivable failure modes for the traffic sign recognition system. For each failure mode, we identify the root causes and examine the resulting effects.

\begin{table}[h]\scriptsize\scriptsize
    \caption{FMEA Worksheet for Traffic Sign Recognition Systems}\label{tab:fmea_worksheet}
    \begin{tabularx}{\textwidth}{@{}p{14mm}p{6mm}p{20mm}p{30mm}p{34mm}}
        \toprule
        Item & Num. & Failure Mode & Cause & Effect \\
        \midrule
        \multirow{8}{16mm}{Traffic Sign Recognition System} 
        & 1 & Erroneously classifying a traffic sign from category A to category B & Defect in the ML algorithm; insufficient training data; ambiguous sign representation & Incorrect recognition may prompt the vehicle to react inappropriately (e.g., abrupt braking or acceleration) based on the misclassified categories. \\
        & 2 & Failure to produce any output & ML algorithm error; software malfunction & The vehicle might not respond to crucial traffic signs, potentially leading to hazardous driving due to the lack of recognition. \\
        & 3 & Delay in real-time processing & Excessive computational demand; suboptimal algorithms; hardware constraints & Slower response times can increase the risk of accidents, such as rear-end collisions, due to delayed vehicle actions. \\
        & 4 & Hardware failure & Sensor malfunctions; power supply issues; physical damage & Loss of sensor input or system shutdown hampers traffic sign detection, thereby impairing vehicle response. \\
        & 5 & Cybersecurity breaches & Cyberattacks; unauthorized access; data tampering & Interference or alteration of traffic sign data may yield incorrect outputs, leading to unsafe vehicular maneuvers. \\
        & 6 & Misclassification due to adversarial samples & Inputs specifically engineered to deceive the ML model & Erroneous classification caused by adversarial inputs can result in unsafe vehicle behavior based on false information. \\
        & 7 & Absence of input & Camera blockage; connectivity issues; adverse environmental conditions & Failure to capture input data can render the system unresponsive, affecting vehicle actions. \\
        & 8 & Delayed wake-up signal & Sensor malfunction; bandwidth limitations; system errors; prioritization problems & A lag in system activation may cause the vehicle to overlook critical traffic signs. \\
        \bottomrule
    \end{tabularx}
\end{table}

\subsubsection{Effect analysis}


Our analysis primarily focuses on Failure Mode 1, where a traffic sign is incorrectly classified from Class A to Class B. The severity of this misclassification is contingent upon both the true class and the misidentified class. To address this, we establish severity levels for various misclassification scenarios and propose a statistical approach to compute their probabilities.

\paragraph{Severity}




To accommodate the spectrum of misclassifications between true and predicted classes, we introduce a severity matrix with levels ranging from 0 to 2:
\begin{itemize}
    \item \textbf{Level 0:} Misclassifications that pose no safety risk (for instance, mistaking a "Pedestrian Crossing" sign for a "School Crossing" sign is unlikely to create dangerous conditions).
    \item \textbf{Level 1:} Misclassifications that have minor adverse effects and are unlikely to result in accidents.
    \item \textbf{Level 2:} Misclassifications that may lead to serious accidents (e.g., misidentifying a "Speed Limit 30" sign as a "Speed Limit 80" sign might cause the vehicle to reach unsafe speeds, potentially triggering a critical incident).
\end{itemize}
Entries marked with “-” indicate correct classifications by the system.


In this study, we concentrate on traffic sign classification using the GTSRB and ArTS datasets. We have assigned severity levels to all possible misclassification events. Tables \ref{tab:severity_gtsrb} and \ref{tab:severity_arts} display the severity levels for a subset of representative traffic signs and their misclassification scenarios. The complete severity matrix for the GTSRB and ArTS dataset is provided in the appendix.

\begin{table}[h]\scriptsize
    \centering
    \caption{Severity Levels for Traffic Sign Classification on the GTSRB Dataset}\label{tab:severity_gtsrb}
    \begin{tabular}{@{}lccccc@{}}
        \toprule
        \multirow{2}{20mm}{Ground Truth} & \multicolumn{5}{@{}c@{}}{Prediction} \\ 
        \cmidrule{2-6}
        & Speed limit 60 & Speed limit 120 & Stop & Danger & Go right \\
        \midrule
        Speed limit 60 & - & \cellcolor{RED!50}2 & \cellcolor{RED!50}2 & \cellcolor{YELLOW!50}1 & \cellcolor{YELLOW!50}1 \\
        Speed limit 120 & \cellcolor{RED!50}2 & - & \cellcolor{RED!50}2 & \cellcolor{YELLOW!50}1 & \cellcolor{YELLOW!50}1 \\
        Stop& \cellcolor{RED!50}2 & \cellcolor{RED!50}2 & - & 0 & 0 \\
        Danger& \cellcolor{YELLOW!50}1 & \cellcolor{YELLOW!50}1 & 0 & - & 0 \\
        Go right& 0 & 0 & 0 & 0 & - \\
        \bottomrule
    \end{tabular}
\end{table}

\begin{table}[h]\scriptsize
    \centering
    \caption{Severity Levels for Traffic Sign Classification on the ArTS Dataset}\label{tab:severity_arts}
    \begin{tabular}{@{}lccccc@{}}
        \toprule
        \multirow{2}{20mm}{Ground Truth} & \multicolumn{5}{@{}c@{}}{Prediction} \\ 
        \cmidrule{2-6}
        & Front or Right & Front or Left & Stop & Speed 60 & Speed 100 \\
        \midrule
        Front or Right & - & \cellcolor{RED!50}2 & \cellcolor{YELLOW!50}1 & \cellcolor{YELLOW!50}1 & \cellcolor{YELLOW!50}1 \\
        Front or Left & \cellcolor{RED!50}2 & - & \cellcolor{YELLOW!50}1 & \cellcolor{YELLOW!50}1 & \cellcolor{YELLOW!50}1 \\
        Stop & \cellcolor{YELLOW!50}1 & \cellcolor{YELLOW!50}1 & - & \cellcolor{RED!50}2 & \cellcolor{RED!50}2 \\
        Speed 60 & \cellcolor{YELLOW!50}1 & \cellcolor{YELLOW!50}1 & \cellcolor{RED!50}2 & - & \cellcolor{RED!50}2 \\
        Speed 100 & \cellcolor{YELLOW!50}1 & \cellcolor{YELLOW!50}1 & \cellcolor{RED!50}2 & \cellcolor{RED!50}2 & - \\
        \bottomrule
    \end{tabular}
\end{table}

\paragraph{Probability}




We estimate misclassification probabilities by constructing a probability matrix derived from experimental data. In our study, ML models were evaluated on the GTSRB test set, with the outcomes recorded in a confusion matrix $C$. The matrix $C$ is an $n \times n$ array, where $n$ is the number of classes; each element $C_{ij}$ counts the number of instances of true class $i$ that were predicted as class $j$. From $C$, we derive a probability matrix $P$, where each element $P_{ij}$ denotes the likelihood of an instance of class $i$ being classified as class $j$ (with $i \neq j$ indicating a misclassification):

\begin{equation}
    P_{ij} = \frac{C_{ij}}{\sum_{i=1}^n \sum_{j=1}^n C_{ij}}
\end{equation}

\subsubsection{Risk Calculation}


We allocate severity scores to the defined severity levels ($sl$) in the severity matrix $S$, where higher scores reflect a greater negative impact on safety. The element $S_{ij}$ is defined as:

\begin{equation}
    S_{ij} =
\begin{cases} 
\sigma(sl), & i \neq j \\
0, & i = j
\end{cases}
\end{equation}



with $\sigma(sl)$ being a monotonically increasing function that maps severity levels to numerical scores.

To determine the overall risk, we perform element-wise multiplication between the severity matrix $S$ and the probability matrix $P$, resulting in the risk matrix $R$:

\begin{equation}
    R = S \odot P
\end{equation}




Here, $\odot$ represents element-wise multiplication. The cumulative risk score, which quantifies the total risk associated with the ML model and system, is then computed by summing all elements of $R$:

\begin{equation}
    \text{Risk Score} = \sum_{i=1}^n \sum_{j=1}^n R_{ij}
\end{equation}


The risk score ranges from 0 to $\sigma(2)$, where $\sigma(2)$ corresponds to the highest severity level.

\subsection{Safety score}



The computed risk score is used to derive a safety score that evaluates the safety of both individual ML models and the overall NVML system. A higher safety score signifies a more secure traffic sign recognition system. The safety score is calculated as follows:

\begin{equation}
    \text{Safety Score} = \frac{1}{1 + \text{Risk Score}}
\end{equation}


This score falls within the interval $(0, 1]$, with values approaching 1 indicating higher safety.

\subsection{Safety-aware weight}


Assume the NVML system employs $T$ independent ML models. By assessing the risks associated with each model, we derive a set of risk scores, $\{rs_1, rs_2, \dots, rs_T\}$. These scores are then used to assign a safety-aware weight to each model, which is applied in both weighted voting and weighted soft voting. The weight for the $i$-th model is determined by:


\begin{equation}
    w_i = \frac{\sum_{j=1}^T rs_j}{rs_i}
\end{equation}


This formulation ensures that models with higher risk scores receive lower weights, thereby prioritizing the influence of safer models and enhancing the overall safety of the decision-making process.

\section{Experiment}\label{sec:experiment}





An experimental study was conducted to assess the effectiveness of different voting strategies within NVML systems applied to traffic sign recognition. The performance metrics considered include classification accuracy (Acc), the computed safety score (Safety), and the counts of misclassifications at severity levels 2 (SL2) and 1 (SL1).

We developed a three-variant traffic sign recognition system that integrates three distinct ML architectures: AlexNet, VGG16, and EfficientNetB0. These models were chosen for their ability to deliver high accuracy while keeping computational demands low. Each network was trained independently on the same training set drawn from the GTSRB dataset.

To derive safety-informed weights, the GTSRB test set—comprising 12,630 images—was evenly split into two portions. The first portion, termed the “weight assignment set,” was used to compute the weights, while the second portion, called the “evaluation set,” was utilized to benchmark the performance of both the individual ML models and the overall three-variant system.

In this experiment, the severity scores are defined as follows:

\begin{equation}
    \sigma(0) = 0, \, \sigma(1) = 1, \, \sigma(2) = 10
\end{equation}


This scoring scheme is designed to impose a heavier penalty on misclassifications at severity level 2, reflecting their more significant impact on safety.

\subsection{Safety-aware Weight Assignment for NVML System}

\subsubsection{Safety score method}


The models were first evaluated using the weight assignment set to obtain their respective safety scores. These scores were then used to calculate safety-aware weights that were allocated to each model. Table \ref{tab:weight_assignment} presents the prediction outcomes and the normalized weights for each model. As anticipated, EfficientNetB0, with the top safety score of $98.14\%$, was assigned the highest weight of $0.573$ in the weighted voting scheme. VGG16, with the next highest safety score, was allocated a weight of $0.301$, while AlexNet, having the lowest safety score, received a weight of $0.126$.

\begin{table}[h]\scriptsize
    \centering
    \caption{Model Prediction Outcomes and Corresponding Weights}\label{tab:weight_assignment}
    \begin{tabular}{@{}lllll@{}}
        \toprule
        Model & Acc (\%) & Safety (\%) & Safety-socre Weight & LLM Weight \\
        \midrule
        AlexNet & 94.28 & 92.02 & 0.126 & 0.206 \\ 
        VGG16 & 97.64 & 96.52 & 0.301 & 0.375 \\ 
        EfficientNetB0 & 98.53 & 98.14 & 0.573 & 0.419 \\
        \bottomrule
    \end{tabular}
\end{table}

\subsubsection{Large language model method}




Recent advances in LLMs have proven beneficial for complex tasks and decision-making processes. Taking advantage of this capability, we employed an LLM—specifically GPT-4o—to generate a set of benchmark weights for our system, which we then compared to the weights obtained via the safety score method.

To guide the LLM, we crafted a specialized prompt containing keywords pertinent to the NVML system and the weight determination criteria. In addition, we provided the LLM with both the severity matrix and the confusion matrix (extracted from predictions on the weight assignment set) in CSV format.

The prompt used during our experiments is shown below:

\begin{table}[h]\scriptsize\scriptsize
\begin{tabular}{@{}p{0.9\textwidth}@{}}
    \toprule
    "Determine the weights for models in the three-version machine learning system that consists of AlexNet, VGG16, and EfficientNetB0—the confusion matrix results from each model tested on the same evaluation dataset. The severity matrix represents the severity of every misclassification. The value in the confusion and severity matrix should be integers. The model produces misclassification with higher severity and probability should be assigned a lower weight. The weight value is in the range of 0 to 1."\\
    \bottomrule
\end{tabular}
\end{table}


The weights produced by the LLM, as listed in the final column of Table \ref{tab:weight_assignment}, reveal that the LLM-based approach yields weights that are more evenly distributed across the models. In contrast, the safety score method assigns weights that more distinctly favor models with superior safety performance.

\subsection{Safety evaluation}

\subsubsection{Individual ML models}


The individual ML models were evaluated using the evaluation set, with the results summarized in Table \ref{tab:gtsrb_safety_ml}. Among the three models, EfficientNetB0 demonstrated superior performance across all metrics, achieving the highest accuracy and safety score, along with the fewest misclassifications at both SL2 and SL1.

\begin{table}[h]\scriptsize
    \centering
    \caption{Performance of Individual ML Models}\label{tab:gtsrb_safety_ml}
    \begin{tabular}{@{}lllll@{}}
        \toprule
        Model & Acc (\%) & Safety (\%) & SL2 & SL1 \\
        \midrule
        AlexNet & 94.17 & 91.26 & 45 & 155 \\ 
        VGG16 & 97.37 & 96.02 & 20 & 62 \\ 
        EfficientNetB0 & \textbf{98.13} & \textbf{96.98} & \textbf{17} & \textbf{27} \\
        \bottomrule
    \end{tabular}
\end{table}



These results indicate a strong correlation between safety score and the frequency of severe misclassifications. Models with fewer high-severity errors tend to achieve higher safety scores; for example, AlexNet, which recorded the highest SL2 misclassification count, correspondingly obtained the lowest safety score among the evaluated models. This emphasizes the need to minimize severe misclassification errors to enhance both system safety and overall performance.

\subsubsection{NVML systems}


The NVML system was further assessed by incorporating various voting mechanisms into the three-variant traffic sign recognition system. Each model processed the same inputs from the evaluation set independently, and their outputs were then combined using different voting methods to produce the final prediction. In scenarios where majority voting failed to yield an output—such as when all three models disagreed—the prediction from AlexNet was used as the default. Table \ref{tab:gtsrb_safety_nvml} details the performance metrics for each voting approach.

\begin{table}[h]\scriptsize
    \centering
    \caption{Performance Metrics for Different Voting Mechanisms}\label{tab:gtsrb_safety_nvml}
    \begin{tabular}{@{}lllll@{}}
        \toprule
        Model & Acc (\%) & Safety (\%) & SL2 & SL1 \\
        \midrule
        Majority & 97.88 & 96.69 & 17 & 56 \\ 
        Weighted & 97.13 & 96.96 & 16 & 27 \\ 
        Simple Soft & 98.32 & 97.12 & 14 & 47 \\ 
        Weighted Soft & 98.40 & \textbf{97.92} & \textbf{11} & \textbf{24} \\ 
        LLM Weighted & 98.24 & 97.05 & 15 & 42 \\ 
        LLM Weighted Soft & \textbf{98.50} & 97.35 & 13 & 42 \\ 
        \bottomrule
    \end{tabular}
\end{table}




The data indicate that the weighted soft voting method outperforms the other strategies, delivering the highest safety score and accuracy while minimizing severe misclassifications. Remarkably, the NVML system’s performance using this method even exceeds that of the best individual model, EfficientNetB0. This result suggests that the NVML system is capable of maintaining robust performance even when incorporating models with lower individual accuracy, such as AlexNet.

Moreover, systems employing soft voting, weighted soft voting, LLM-derived weight voting, and our proposed weighted soft voting all yield better accuracy than the top-performing single model. This further supports the effectiveness of our weight assignment approach in boosting both safety and classification accuracy.

While the LLM-based weighted soft voting method improves safety relative to majority and simple soft voting, its enhancements are less pronounced than those achieved by our proposed safety score-based method. Despite testing multiple prompt variations using the same confusion and severity matrices, the results shown in Table \ref{tab:gtsrb_safety_nvml} represent the optimal outcomes. Future efforts may involve refining the guidance provided to the LLM to further optimize the derivation of safety-aware weights.

\subsection{Adversarial Robustness}

Adversarial attacks present a critical challenge to the reliability of ML systems, especially in safety-critical domains. These attacks exploit model vulnerabilities by introducing imperceptible perturbations to input data, leading to incorrect predictions that can have severe consequences \citep{goodfellow2014explaining}. Traditional single-model systems often fail to provide sufficient robustness against such attacks \citep{papernot2017practical}.

The NVML systems leverage model redundancy and diversity to enhance robustness under adversarial conditions. By integrating multiple independently trained models, NVML systems reduce the likelihood of simultaneous failures across all models, significantly mitigating the effects of adversarial inputs \citep{biggio2018wild}. This approach not only improves system reliability but also aligns with the need for higher safety standards in adversarial environments.

This study begins by evaluating individual ML models' performance under adversarial attacks to establish a baseline for vulnerability. It then evaluates an NVML system's robustness, focusing on how redundancy and diversity can mitigate the vulnerabilities observed in single-model systems. The results will provide insights into the effectiveness of NVML systems in real-world adversarial scenarios and guide the design of more resilient ML applications.

In this evaluation, we address the following research questions.

\begin{itemize}
    \item RQ1: How do individual ML models perform under different types of adversarial attacks?
    \item RQ2: How are NVML systems tolerant to adversarial attacks?
\end{itemize}

In this study, adversarial samples were crafted using the FGSM and PGD methods, both widely recognized approaches for generating adversarial perturbations. The perturbation magnitude $\epsilon$ was set to $0.03$ to achieve a balance between input distortion and attack effectiveness, while a stronger attack was simulated with $\epsilon = 0.05$ to highlight the impact of increased perturbation on model performance. The hyperparameters corresponding to the $\epsilon$ values used for PGD attacks are detailed in Table \ref{tab:pgd_hyper}. Similar to the safety evaluation process, the GTSRB and ArTS test set was randomly divided into two equal-sized subsets: a "weight assignment set" and an "evaluation set." These subsets were newly generated and are distinct from the previously used sets. These samples were derived from the new "evaluation set" and introduced small perturbations to the input images while increasing the likelihood of incorrect predictions.

\begin{table}[h]\scriptsize
    \centering
    \caption{PGD hyperparameters}\label{tab:pgd_hyper}
    \begin{tabular}{@{}lll@{}}
        \toprule
        Hyperparameter & Value 1 & Value 2\\
        \midrule
        $\epsilon$ & 0.03 & 0.05 \\
        $\alpha$ & 0.006 & 0.01  \\
        Number of Iterations & 10 & 10  \\
        Random Start & Ture & True  \\
        \bottomrule
    \end{tabular}
\end{table}

The evaluation metrics include accuracy (Acc), robustness accuracy (RAcc), safety score (Safety), robustness safety score (RSafety), the count of SL2 and SL1 misclassifications, and the robustness counts for SL2 and SL1 misclassifications (RSL2 and RSL1).

\subsubsection{Adversarial robustness of ML model}

To evaluate the robustness of AlexNet, VGG16, and EfficientNetB0 against adversarial perturbations, adversarial datasets were generated using both FGSM and PGD attack methods at different $\epsilon$ values from the GTSRB and ArTS evaluation sets. Performance on the original datasets is summarized in Tables \ref{tbl:orm} (GTSRB) and \ref{tbl:arts_model_original} (ArTS), while Tables \ref{tbl:arm}, \ref{tbl:arm5}, \ref{tbl:armp3}, and \ref{tbl:armp} present the adversarial results for GTSRB under FGSM and PGD attacks. Similarly, Tables \ref{tbl:arts_model_fgsm_3} and \ref{tbl:arts_model_fgsm_5} show the adversarial outcomes for the ArTS dataset under FGSM. In these tables, the values in parentheses indicate the differences from the original results.

\begin{table}[H]\scriptsize
    \centering
    \caption{Original GTSRB set results - model}\label{tbl:orm}
    \begin{tabular}{@{}lllll@{}}
        \toprule
        Model & Acc (\%) & Safety (\%) & SL2 & SL1 \\
        \midrule
        AlexNet & 94.28 & 92.88 & 33 & 154 \\
        VGG16 & 97.34 & 95.87 & 20 & 72 \\
        EfficientNetB0 & 98.16 & 97.15 & 16 & 25 \\
        \bottomrule
    \end{tabular}
\end{table}

\begin{table}[H]\scriptsize
    \caption{Adversarial GTSRB set results - model (with changes) - FGSM - $\epsilon = 0.03$}\label{tbl:arm}
    \begin{tabular*}{\textwidth}{@{\extracolsep\fill}lllll}
        \toprule
        Model & RAcc (\%) & RSafety (\%) & RSL2 & RSL1 \\
        \midrule
        AlexNet & 92.95 (-1.33) & 89.82 (-3.06) & 52 (+19) & 196 (+42) \\
        VGG16 & 96.50 (-0.84) & 94.27 (-1.60) & 29 (+9) & 94 (+22) \\
        EfficientNetB0 & 97.32 (-0.84) & 95.97 (-1.18) & 22 (+6) & 45 (+20) \\
        \bottomrule
    \end{tabular*}
\end{table}

\begin{table}[H]\scriptsize
    \caption{Adversarial GTSRB set results - model (with changes) - FGSM - $\epsilon = 0.05$}\label{tbl:arm5}
    \begin{tabular*}{\textwidth}{@{\extracolsep\fill}lllll}
        \toprule
        Model & RAcc (\%) & RSafety (\%) & RSL2 & RSL1 \\
        \midrule
        AlexNet & 92.15 (-2.13) & 87.37 (-5.51) & 69 (+36) & 223 (+69) \\
        VGG16 & 96.04 (-1.30) & 93.44 (-2.43) & 33 (+13) & 113 (+41) \\
        EfficientNetB0 & 96.77 (-1.39) & 95.21 (-1.94) & 25 (+9) & 68 (+43) \\
        \bottomrule
    \end{tabular*}
\end{table}

\begin{table}[H]\scriptsize
    \caption{Adversarial GTSRB set results - model (with changes) - PGD - $\epsilon = 0.03$}\label{tbl:armp3}
    \begin{tabular*}{\textwidth}{@{\extracolsep\fill}lllll}
        \toprule
        Model & RAcc (\%) & RSafety (\%) & RSL2 & RSL1 \\
        \midrule
        AlexNet & 46.90 (-47.38) & 45.90 (-46.98) & 539 (+506) & 2053 (+1899) \\
        VGG16 & 56.77 (-40.57) & 64.52 (-31.35) & 195 (+175) & 1522 (+1450) \\
        EfficientNetB0 & 55.47 (-42.69) & 70.23 (-26.92) & 118 (+102) & 1497 (+1472) \\
        \bottomrule
    \end{tabular*}
\end{table}

\begin{table}[H]\scriptsize
    \caption{Adversarial GTSRB set results - model (with changes) - PGD - $\epsilon = 0.05$}\label{tbl:armp}
    \begin{tabular*}{\textwidth}{@{\extracolsep\fill}lllll}
        \toprule
        Model & RAcc (\%) & RSafety (\%) & RSL2 & RSL1 \\
        \midrule
        AlexNet & 41.62 (-52.66) & 42.99 (-49.89) & 614 (+581) & 2236 (+2082) \\
        VGG16 & 51.29 (-46.05) & 57.06 (-38.81) & 310 (+290) & 1652 (+1580) \\
        EfficientNetB0 & 51.12 (-47.04) & 65.58 (-31.57) & 169 (+153) & 1625 (+1600) \\
        \bottomrule
    \end{tabular*}
\end{table}

\begin{table}[H]\scriptsize
    \centering
    \caption{Original ArTS set results - model}\label{tbl:arts_model_original}
    \begin{tabular}{@{}lllll@{}}
        \toprule
        Model & Acc (\%) & Safety (\%) & SL2 & SL1 \\
        \midrule
        AlexNet & 98.55 & 95.13 & 23 & 48 \\
        VGG16 & 99.56 & 99.65 & 0 & 19 \\
        EfficientNetB0 & 99.39 & 99.65 & 0 & 19 \\
        \bottomrule
    \end{tabular}
\end{table}

\begin{table}[h]\scriptsize
    \caption{Adversarial ArTS set results - model (with changes) - FGSM - $\epsilon = 0.03$}\label{tbl:arts_model_fgsm_3}
    \begin{tabular*}{\textwidth}{@{\extracolsep\fill}lllll}
        \toprule
        Model & RAcc (\%) & RSafety (\%) & RSL2 & RSL1 \\
        \midrule
        AlexNet & 97.39 (-1.16) & 93.11 (-2.02) & 33 (+10) & 72 (+24) \\
        VGG16 & 98.55 (-1.01) & 98.98 (-0.67) & 1 (+1) & 46 (+27) \\
        EfficientNetB0 & 99.25 (-0.14) & 99.58 (-0.07) & 0 (0) & 23 (+4) \\
        \bottomrule
    \end{tabular*}
\end{table}

\begin{table}[h]\scriptsize
    \caption{Adversarial ArTS set results - model (with changes) - FGSM - $\epsilon = 0.05$}\label{tbl:arts_model_fgsm_5}
    \begin{tabular*}{\textwidth}{@{\extracolsep\fill}lllll}
        \toprule
        Model & RAcc (\%) & RSafety (\%) & RSL2 & RSL1 \\
        \midrule
        AlexNet & 95.45 (-3.10) & 90.34 (-4.79) & 44 (+21) & 141 (+93) \\
        VGG16 & 97.39 (-2.17) & 97.61 (-2.04) & 5 (+5) & 83 (+64) \\
        EfficientNetB0 & 99.01 (-0.38) & 99.16 (-0.49) & 1 (+1) & 36 (+17) \\
        \bottomrule
    \end{tabular*}
\end{table}

\begin{table}[h]\scriptsize
    \caption{Adversarial ArTS set results - model (with changes) - PGD - $\epsilon = 0.03$}\label{tbl:arts_model_pgd_3}
    \begin{tabular*}{\textwidth}{@{\extracolsep\fill}lllll}
        \toprule
        Model & RAcc (\%) & RSafety (\%) & RSL2 & RSL1 \\
        \midrule
        AlexNet & 82.11 (-16.44) & 82.33 (-12.80) & 69 (+46) & 479 (+431) \\
        VGG16 & 90.24 (-9.32) & 91.60 (-8.05) & 26 (+26) & 238 (+219) \\
        EfficientNetB0 & 92.14 (-7.25) & 89.76 (-9.89) & 40 (+40) & 220 (+201) \\
        \bottomrule
    \end{tabular*}
\end{table}

\begin{table}[h]\scriptsize
    \caption{Adversarial ArTS set results - model (with changes) - PGD - $\epsilon = 0.05$}\label{tbl:arts_model_pgd_5}
    \begin{tabular*}{\textwidth}{@{\extracolsep\fill}lllll}
        \toprule
        Model & RAcc (\%) & RSafety (\%) & RSL2 & RSL1 \\
        \midrule
        AlexNet & 77.86 (-20.69) & 80.26 (-14.87) & 71 (+48) & 626 (+578) \\
        VGG16 & 89.40 (-10.16) & 90.85 (-8.80) & 29 (+29) & 257 (+238) \\
        EfficientNetB0 & 92.21 (-7.18) & 89.58 (-10.07) & 42 (+42) & 212 (+193) \\
        \bottomrule
    \end{tabular*}
\end{table}

For FGSM attacks at $\epsilon = 0.03$ on the GTSRB dataset (Table \ref{tbl:arm}), all three models show only modest performance declines. For example, AlexNet’s accuracy decreases by 1.33\% (from 94.28\% to 92.95\%), and its safety score drops by 3.06\%, while SL2 and SL1 misclassifications increase by 19 and 42, respectively. VGG16 and EfficientNetB0 experience even smaller degradations. When the perturbation is increased to $\epsilon = 0.05$ (Table \ref{tbl:arm5}), the drops become slightly more pronounced—for instance, AlexNet’s accuracy falls by 2.13\% and its safety score by 5.51\%, with corresponding increases in SL2 and SL1 misclassifications.

In stark contrast, the iterative PGD attack has a far more severe impact. Under PGD with $\epsilon = 0.03$ (Table \ref{tbl:armp3}), AlexNet’s accuracy plummets by 47.38\%, and its safety score drops by 46.98\%, accompanied by a massive surge in SL2 (an increase of 506) and SL1 misclassifications (an increase of 1899). Similar significant degradations are observed for VGG16 and EfficientNetB0. At a higher perturbation level of $\epsilon = 0.05$ (Table \ref{tbl:armp}), all models suffer even larger drops in accuracy (over 40–50\%) and safety scores, with dramatic increases in misclassification counts.

For the ArTS dataset under FGSM attacks, as shown in Tables \ref{tbl:arts_model_fgsm_3} and \ref{tbl:arts_model_fgsm_5}, the performance degradation is less severe. At $\epsilon = 0.03$, the decreases in accuracy and safety scores are minimal (e.g., AlexNet drops by only 1.16\% in accuracy), with slight increases in misclassifications. However, at $\epsilon = 0.05$, the negative impact becomes more apparent; for instance, AlexNet’s accuracy declines by 3.10\% and misclassifications increase accordingly.

In summary, while the models demonstrate relative robustness to FGSM attacks—especially at lower $\epsilon$ levels—they are highly susceptible to iterative PGD attacks, which cause substantial reductions in accuracy and safety scores along with significant increases in misclassification rates. These findings emphasize the critical need for enhanced adversarial defense strategies in safety-critical applications like traffic sign recognition.

\textbf{Answer to RQ1:} The experimental results reveal that the models withstand FGSM attacks with only slight performance degradation; however, they are considerably vulnerable to PGD attacks, which lead to severe drops in both accuracy and safety scores and a marked rise in misclassifications.

\subsubsection{Adversarial robustness of NVML system}

To assess the robustness of our proposed three-model NVML system, we evaluated its performance on both the original and adversarial versions of the GTSRB and ArTS datasets. In our experiments, the system—which integrates AlexNet, VGG16, and EfficientNetB0—was compared against a single-model baseline (AlexNet) under various adversarial attacks, including one-step attacks (FGSM) and iterative attacks (PGD) at different perturbation strengths.

\begin{table}[h]\scriptsize
    \centering
    \caption{Original GTSRB set results - NVML}\label{tbl:orn}
    \begin{tabular}{@{}lllll@{}}
        \toprule
        Voting Mechanism & Acc (\%) & Safety (\%) & SL2 & SL1 \\
        \midrule
        AlexNet & 94.28 & 92.88 & 33 & 154 \\
        Majority & 97.89 & 96.63 & 16 & 60 \\
        Weighted & 98.16 & 97.15 & 16 & 25 \\
        Simple Sot & 98.34 & 96.90 & 15 & 52 \\
        Weighted Soft & 98.35 & 97.41 & 14 & 28 \\
        LLM Weighted & 98.29 & 97.14 & 14 & 46 \\
        LLM Weighted Soft & 98.61 & 97.15 & 14 & 45 \\
        \bottomrule
    \end{tabular}
\end{table}

\begin{table}[h]\scriptsize
    \caption{Adversarial GTSRB set results - NVML (with changes) - FGSM - $\epsilon = 0.03$}\label{tbl:arn}
    \begin{tabular*}{\textwidth}{@{\extracolsep\fill}lllll}
        \toprule
        Voting Mechanism & RAcc (\%) & RSafety (\%) & RSL2 & RSL1 \\
        \midrule
        AlexNet & 92.95 (-1.33) & 89.82 (-3.06) & 52 (+19) & 196 (+42) \\
        Majority & 97.66 (-0.23) & 96.68 (+0.05) & 15 (-1) & 67 (+7) \\
        Weighted & 97.82 (-0.35) & 96.24 (-0.91) & 22 (+6) & 27 (+2) \\
        Simple Soft & 98.07 (-0.27) & 97.08 (+0.18) & 13 (-2) & 60 (+8) \\
        Weighted Soft & 98.15 (-0.21) & 97.29 (-0.12) & 15 (+1) & 26 (-2) \\
        LLM Weighted & 98.10 (-0.19) & 96.99 (-0.15) & 15 (+1) & 46 (0) \\
        LLM Weighted Soft & 98.40 (-0.21) & 97.11 (-0.04) & 14 (0) & 48 (+3) \\
        \bottomrule
    \end{tabular*}
\end{table}

\begin{table}[h]\scriptsize
    \caption{Adversarial GTSRB set results - NVML (with changes) - FGSM - $\epsilon = 0.05$}\label{tbl:arn5}
    \begin{tabular*}{\textwidth}{@{\extracolsep\fill}lllll}
        \toprule
        Voting Mechanism & RAcc (\%) & RSafety (\%) & RSL2 & RSL1 \\
        \midrule
        AlexNet & 92.15 (-2.13) & 87.37 (-5.51) & 69 (+36) & 223 (+69) \\
        Majority & 97.57 (-0.32) & 96.78 (+0.15) & 14 (-2) & 70 (+10) \\
        Weighted & 97.78 (-0.38) & 96.33 (-0.82) & 21 (+5) & 30 (+5) \\
        Simple Soft & 98.03 (-0.31) & 97.02 (+0.12) & 13 (-2) & 64 (+12) \\
        Weighted Soft & 98.13 (-0.22) & 97.54 (+0.13) & 13 (-1) & 29 (+1) \\
        LLM Weighted & 98.05 (-0.24) & 97.10 (-0.04) & 14 (0) & 48 (+2) \\
        LLM Weighted Soft & 98.33 (-0.28) & 97.06 (-0.09) & 14 (0) & 51 (+6) \\
        \bottomrule
    \end{tabular*}
\end{table}

\begin{table}[H]\scriptsize
    \caption{Adversarial GTSRB set results - NVML (with changes) - PGD - $\epsilon = 0.03$}\label{tbl:arnp3}
    \begin{tabular*}{\textwidth}{@{\extracolsep\fill}lllll}
        \toprule
        Voting Mechanism & RAcc (\%) & RSafety (\%) & RSL2 & RSL1 \\
        \midrule
        AlexNet & 46.90 (-47.38) & 45.90 (-46.98) & 539 (+506) & 2053 (+1899) \\
        Majority & 61.74 (-36.15) & 55.68 (-40.95) & 338 (+322) & 1646 (+1586) \\
        Weighted & 60.74 (-37.42) & 76.29 (-20.86) & 62 (+46) & 1343 (+1318) \\
        Simple Soft & 61.85 (-36.00) & 75.90 (-21.00) & 66 (+51) & 1345 (+1293) \\
        Weighted Soft & 62.20 (-36.15) & 77.93 (-19.47) & 48 (+34) & 1308 (+1280) \\
        LLM Weighted & 60.90 (-37.39) & 76.10 (-21.04) & 63 (+49) & 1353 (+1307) \\
        LLM Weighted Soft & 62.36 (-36.25) & 76.90 (-20.25) & 58 (+44) & 1317 (+1272) \\
        \bottomrule
    \end{tabular*}
\end{table}

\begin{table}[H]\scriptsize
    \caption{Adversarial GTSRB set results - NVML (with changes) - PGD - $\epsilon = 0.05$}\label{tbl:arnp}
    \begin{tabular*}{\textwidth}{@{\extracolsep\fill}lllll}
        \toprule
        Voting Mechanism & RAcc (\%) & RSafety (\%) & RSL2 & RSL1 \\
        \midrule
        AlexNet & 41.62 (-52.66) & 42.99 (-49.89) & 614 (+581) & 2236 (+2082) \\
        Majority & 61.69 (-36.20) & 56.31 (-40.32) & 324 (+308) & 1659 (+1599) \\
        Weighted & 60.86 (-37.30) & 76.15 (-21.00) & 64 (+48) & 1338 (+1313) \\
        Simple Soft & 61.87 (-36.47) & 75.57 (-21.33) & 70 (+55) & 1341 (+1289) \\
        Weighted Soft & 62.17 (-36.18) & 77.88 (-19.53) & 48 (+34) & 1314 (+1286) \\
        LLM Weighted & 60.93 (-37.36) & 75.93 (-21.21) & 66 (+52) & 1342 (+1296) \\
        LLM Weighted Soft & 62.42 (-36.19) & 77.25 (-19.90) & 54 (+40) & 1320 (+1275) \\
        \bottomrule
    \end{tabular*}
\end{table}

\begin{table}[H]\scriptsize
    \centering
    \caption{Original ArTS set results - NVML}\label{tbl:arts_nvml_original}
    \begin{tabular}{@{}lllll@{}}
        \toprule
        Voting Mechanism & Acc (\%) & Safety (\%) & SL2 & SL1 \\
        \midrule
        AlexNet & 98.55 & 95.13 & 23 & 48 \\
        Majority & 99.57 & 99.78 & 0 & 12 \\
        Weighted & 99.39 & 99.65 & 0 & 19 \\
        Simple Sot & 99.59 & 99.63 & 0 & 20 \\
        Weighted Soft & 99.65 & 99.68 & 0 & 17 \\
        LLM Weighted & 99.55 & 99.78 & 0 & 12 \\
        LLM Weighted Soft & 99.57 & 99.61 & 0 & 21 \\
        \bottomrule
    \end{tabular}
\end{table}

\begin{table}[h]\scriptsize
    \caption{Adversarial ArTS set results - NVML (with changes) - FGSM - $\epsilon = 0.03$}\label{tbl:arts_nvml_fgsm_3}
    \begin{tabular*}{\textwidth}{@{\extracolsep\fill}lllll}
        \toprule
        Voting Mechanism & RAcc (\%) & RSafety (\%) & RSL2 & RSL1 \\
        \midrule
        AlexNet & 97.39 (-1.16) & 93.11 (-2.02) & 33 (+10) & 72 (+24) \\
        Majority & 99.31 (-0.26) & 99.65 (-0.13) & 0 (0) & 19 (+7) \\
        Weighted & 99.20 (-0.19) & 99.57 (-0.08) & 0 (0) & 23 (+4) \\
        Simple Soft & 99.44 (-0.15) & 99.56 (-0.07) & 0 (0) & 24 (+4) \\
        Weighted Soft & 99.55 (-0.10) & 99.61 (-0.07) & 0 (0) & 21 (+4) \\
        LLM Weighted & 99.31 (-0.24) & 99.65 (-0.13) & 0 (0) & 19 (+7) \\
        LLM Weighted Soft & 99.46 (-0.11) & 99.57 (-0.04) & 0 (0) & 23 (+2) \\
        \bottomrule
    \end{tabular*}
\end{table}

\begin{table}[h]\scriptsize
    \caption{Adversarial ArTS set results - NVML (with changes) - FGSM - $\epsilon = 0.05$}\label{tbl:arts_nvml_fgsm_5}
    \begin{tabular*}{\textwidth}{@{\extracolsep\fill}lllll}
        \toprule
        Voting Mechanism & RAcc (\%) & RSafety (\%) & RSL2 & RSL1 \\
        \midrule
        AlexNet & 95.45 (-3.10) & 90.34 (-4.79) & 44 (+21) & 141 (+93) \\
        Majority & 99.20 (-0.37) & 99.25 (-0.53) & 1 (+1) & 31 (+19) \\
        Weighted & 99.08 (-0.31) & 99.23 (-0.42) & 1 (+1) & 32 (+13) \\
        Simple Soft & 99.31 (-0.28) & 99.28 (-0.35) & 1 (+1) & 29 (+9) \\
        Weighted Soft & 99.54 (-0.11) & 99.57 (-0.11) & 0 (0) & 23 (+6) \\
        LLM Weighted & 99.19 (-0.36) & 99.26 (-0.52) & 1 (+1) & 30 (+18) \\
        LLM Weighted Soft & 99.35 (-0.22) & 99.45 (-0.16) & 0 (0) & 30 (+9) \\
        \bottomrule
    \end{tabular*}
\end{table}

\begin{table}[h]\scriptsize
    \caption{Adversarial ArTS set results - NVML (with changes) - PGD - $\epsilon = 0.03$}\label{tbl:arts_nvml_pgd_3}
    \begin{tabular*}{\textwidth}{@{\extracolsep\fill}lllll}
        \toprule
        Voting Mechanism & RAcc (\%) & RSafety (\%) & RSL2 & RSL1 \\
        \midrule
        AlexNet & 82.11 (-16.44) & 82.33 (-12.80) & 69 (+46) & 479 (+431) \\
        Majority & 90.98 (-8.59) & 89.58 (-10.20) & 43 (+43) & 202 (+190) \\
        Weighted & 91.69 (-7.70) & 88.38 (-11.27) & 49 (+49) & 224 (+205) \\
        Simple Soft & 92.10 (-7.49) & 90.64 (-8.99) & 38 (+38) & 181 (+161) \\
        Weighted Soft & 93.07 (-6.58) & 94.09 (-5.59) & 15 (+15) & 191 (+174) \\
        LLM Weighted & 91.53 (-8.02) & 89.53 (-10.25) & 42 (+42) & 215 (+203) \\
        LLM Weighted Soft & 92.45 (-7.12) & 92.06 (-7.55) & 27 (+27) & 198 (+177) \\
        \bottomrule
    \end{tabular*}
\end{table}

\begin{table}[h]\scriptsize
    \caption{Adversarial ArTS set results - NVML (with changes) - PGD - $\epsilon = 0.05$}\label{tbl:arts_nvml_pgd_5}
    \begin{tabular*}{\textwidth}{@{\extracolsep\fill}lllll}
        \toprule
        Voting Mechanism & RAcc (\%) & RSafety (\%) & RSL2 & RSL1 \\
        \midrule
        AlexNet & 77.86 (-20.69) & 80.26 (-14.87) & 71 (+48) & 626 (+578) \\
        Majority & 90.57 (-9.00) & 89.06 (-10.72) & 46 (+46) & 207 (+195) \\
        Weighted & 91.64 (-7.75) & 88.25 (-11.40) & 50 (+50) & 223 (+204) \\
        Simple Soft & 91.73 (-7.86) & 90.27 (-9.36) & 39 (+39) & 195 (+175) \\
        Weighted Soft & 92.82 (-6.83) & 94.07 (-5.61) & 14 (+14) & 202 (+185) \\
        LLM Weighted & 91.12 (-8.43) & 88.83 (-10.95) & 46 (+46) & 223 (+211) \\
        LLM Weighted Soft & 92.23 (-7.34) & 91.69 (-7.92) & 29 (+29) & 202 (+181) \\
        \bottomrule
    \end{tabular*}
\end{table}

As shown in Table \ref{tbl:orn}, the original GTSRB results indicate that while AlexNet alone achieves decent performance, the NVML system, through various voting mechanisms, delivers improved accuracy and safety. When the system was subjected to FGSM attacks (Tables \ref{tbl:arn} and \ref{tbl:arn5}), the performance degradation was noticeably lower in the NVML configurations compared to the single-model baseline. In particular, although all voting strategies experienced some decline under increasing perturbation levels, the reduction in accuracy and safety was substantially less pronounced for the NVML system.

The advantage of the NVML system becomes even more evident under stronger iterative attacks such as PGD. Tables \ref{tbl:arnp3} and \ref{tbl:arnp} demonstrate that the single-model approach (AlexNet) suffers from dramatic drops in both accuracy and safety, as well as significant increases in severe misclassifications. In contrast, the NVML system maintains considerably higher performance across all metrics. Notably, among the various voting mechanisms evaluated, weighted soft voting consistently exhibits the best robustness—preserving the highest safety scores and the smallest increases in severe misclassifications.

Similar trends are observed in the experiments on the ArTS dataset (Tables \ref{tbl:arts_nvml_original}, \ref{tbl:arts_nvml_fgsm_3}, and \ref{tbl:arts_nvml_fgsm_5}), where the NVML system again outperforms the single-model baseline under adversarial conditions. The overall experimental results indicate that model redundancy and the use of fine-grained voting mechanisms, particularly weighted soft voting, significantly enhance the system’s ability to withstand adversarial perturbations.

In summary, these findings underscore the significant advantage of NVML systems in adversarial environments. By leveraging multiple models and employing advanced voting strategies—especially weighted soft voting—the NVML system maintains high accuracy and safety, effectively mitigating the impact of both weak and strong adversarial attacks.

\textbf{Answer to RQ2:} The proposed NVML system demonstrates markedly enhanced robustness compared to a single-model approach. In particular, weighted soft voting consistently achieves superior performance, maintaining high accuracy and safety while effectively reducing severe misclassifications under adversarial perturbations. These results highlight the critical benefit of model redundancy and fine-grained voting mechanisms for safety-critical applications such as traffic sign recognition.

\section{Conclusion}\label{sec:conclusion}

In this study, we investigated various voting strategies within an NVML system designed for safety-critical traffic sign recognition. By developing a safety metric grounded in the FMEA method, we were able to assess system safety and assign safety-aware weights for both weighted voting and weighted soft voting approaches. Our experimental results clearly indicate that safety-aware weighted soft voting not only improves overall accuracy and safety scores but also robustly reduces severe misclassification instances.

Most notably, the NVML system that leverages model redundancy combined with safety-aware weighted soft voting exhibits exceptional robustness under adversarial attacks—even when subjected to strong perturbations such as PGD. While our exploration of LLM-based weight assignment revealed some potential improvements over traditional majority and soft voting, its performance gains were modest compared to those achieved through weighted soft voting.

Overall, our findings demonstrate that incorporating multiple model versions and employing safety-aware weighted soft voting significantly enhances the reliability and resilience of traffic sign recognition systems. This approach provides a promising pathway for improving the robustness of ML systems in autonomous driving applications.

Future work could expand the NVML system to include a broader range of models, further enhancing robustness through increased diversity, and extend the framework to other ML tasks to deepen our understanding of reliability and safety in real-world scenarios.

\section*{Acknowledgements}
  This work was supported by JST SPRING Grant Number JPMJSP2124, and partly supported by JSPS KAKENHI Grant Number 22K17871.

\bibliographystyle{elsarticle-num-names}

\begin{appendices}

\section{Complete severity matrix of GTSRB and ArTS}

We define the severity levels for all possible misclassifications of traffic signs in the GTSRB and ArTS dataset. 
In Table \ref{Complete Severity Matrix of GTSRB} and \ref{SecondSeverityMatrix}, the ground truth traffic signs are listed in the first column, while the class number of predicted signs is presented at the top line. The severity level is either 0, 1, or 2, indicating that a higher severity level means a more dangerous misclassification.

\begin{sidewaystable}
\setlength{\tabcolsep}{0pt}
\caption{Severity level definitions for misclassifications of traffic signs in GTSRB}\label{Complete Severity Matrix of GTSRB}
\begin{tabular*}{\textwidth}{@{\extracolsep\fill}lccccccccccccccccccccccccccccccccccccccccccc}
\toprule
\multirow{2}{22mm}{\textbf{Ground Truth}} & \multicolumn{43}{@{}c@{}}{\textbf{Prediction}} \\ 
\cmidrule{2-44}
& \textbf{0} & \textbf{1} & \textbf{2} & \textbf{3} & \textbf{4} & \textbf{5} & \textbf{6} & \textbf{7} & \textbf{8} & \textbf{9} & \textbf{10} & \textbf{11} & \textbf{12} & \textbf{13} & \textbf{14} & \textbf{15} & \textbf{16} & \textbf{17} & \textbf{18} & \textbf{19} & \textbf{20} & \textbf{21} & \textbf{22} & \textbf{23} & \textbf{24} & \textbf{25} & \textbf{26} & \textbf{27} & \textbf{28} & \textbf{29} & \textbf{30} & \textbf{31} & \textbf{32} & \textbf{33} & \textbf{34} & \textbf{35} & \textbf{36} & \textbf{37} & \textbf{38} & \textbf{39} & \textbf{40} & \textbf{41} & \textbf{42} \\
\midrule
\textbf{0: Speed limit 20} & - & \cellcolor{YELLOW!50}1 & \cellcolor{RED!50}2 & \cellcolor{RED!50}2 & \cellcolor{RED!50}2 & \cellcolor{RED!50}2 & \cellcolor{YELLOW!50}1 & \cellcolor{RED!50}2 & \cellcolor{RED!50}2 & \cellcolor{YELLOW!50}1 & \cellcolor{YELLOW!50}1 & \cellcolor{YELLOW!50}1 & \cellcolor{YELLOW!50}1 & \cellcolor{YELLOW!50}1 & \cellcolor{RED!50}2 & \cellcolor{YELLOW!50}1 & \cellcolor{YELLOW!50}1 & \cellcolor{RED!50}2 & \cellcolor{YELLOW!50}1 & \cellcolor{YELLOW!50}1 & \cellcolor{YELLOW!50}1 & \cellcolor{YELLOW!50}1 & \cellcolor{YELLOW!50}1 & \cellcolor{YELLOW!50}1 & \cellcolor{YELLOW!50}1 & \cellcolor{YELLOW!50}1 & \cellcolor{YELLOW!50}1 & \cellcolor{YELLOW!50}1 & \cellcolor{YELLOW!50}1 & \cellcolor{YELLOW!50}1 & \cellcolor{YELLOW!50}1 & \cellcolor{YELLOW!50}1 & \cellcolor{YELLOW!50}1 & \cellcolor{YELLOW!50}1 & \cellcolor{YELLOW!50}1 & \cellcolor{YELLOW!50}1 & \cellcolor{YELLOW!50}1 & \cellcolor{YELLOW!50}1 & \cellcolor{YELLOW!50}1 & \cellcolor{YELLOW!50}1 & \cellcolor{YELLOW!50}1 & \cellcolor{YELLOW!50}1 & \cellcolor{YELLOW!50}1 \\
\textbf{1: Speed limit 30} & \cellcolor{YELLOW!50}1 & - & \cellcolor{RED!50}2 & \cellcolor{RED!50}2 & \cellcolor{RED!50}2 & \cellcolor{RED!50}2 & \cellcolor{YELLOW!50}1 & \cellcolor{RED!50}2 & \cellcolor{RED!50}2 & \cellcolor{YELLOW!50}1 & \cellcolor{YELLOW!50}1 & \cellcolor{YELLOW!50}1 & \cellcolor{YELLOW!50}1 & \cellcolor{YELLOW!50}1 & \cellcolor{RED!50}2 & \cellcolor{YELLOW!50}1 & \cellcolor{YELLOW!50}1 & \cellcolor{RED!50}2 & \cellcolor{YELLOW!50}1 & \cellcolor{YELLOW!50}1 & \cellcolor{YELLOW!50}1 & \cellcolor{YELLOW!50}1 & \cellcolor{YELLOW!50}1 & \cellcolor{YELLOW!50}1 & \cellcolor{YELLOW!50}1 & \cellcolor{YELLOW!50}1 & \cellcolor{YELLOW!50}1 & \cellcolor{YELLOW!50}1 & \cellcolor{YELLOW!50}1 & \cellcolor{YELLOW!50}1 & \cellcolor{YELLOW!50}1 & \cellcolor{YELLOW!50}1 & \cellcolor{YELLOW!50}1 & \cellcolor{YELLOW!50}1 & \cellcolor{YELLOW!50}1 & \cellcolor{YELLOW!50}1 & \cellcolor{YELLOW!50}1 & \cellcolor{YELLOW!50}1 & \cellcolor{YELLOW!50}1 & \cellcolor{YELLOW!50}1 & \cellcolor{YELLOW!50}1 & \cellcolor{YELLOW!50}1 & \cellcolor{YELLOW!50}1 \\
\textbf{2: Speed limit 50} & \cellcolor{RED!50}2 & \cellcolor{YELLOW!50}1 & - & \cellcolor{YELLOW!50}1 & \cellcolor{RED!50}2 & \cellcolor{RED!50}2 & \cellcolor{YELLOW!50}1 & \cellcolor{RED!50}2 & \cellcolor{RED!50}2 & \cellcolor{YELLOW!50}1 & \cellcolor{YELLOW!50}1 & \cellcolor{YELLOW!50}1 & \cellcolor{YELLOW!50}1 & \cellcolor{YELLOW!50}1 & \cellcolor{RED!50}2 & \cellcolor{YELLOW!50}1 & \cellcolor{YELLOW!50}1 & \cellcolor{RED!50}2 & \cellcolor{YELLOW!50}1 & \cellcolor{YELLOW!50}1 & \cellcolor{YELLOW!50}1 & \cellcolor{YELLOW!50}1 & \cellcolor{YELLOW!50}1 & \cellcolor{YELLOW!50}1 & \cellcolor{YELLOW!50}1 & \cellcolor{YELLOW!50}1 & \cellcolor{YELLOW!50}1 & \cellcolor{YELLOW!50}1 & \cellcolor{YELLOW!50}1 & \cellcolor{YELLOW!50}1 & \cellcolor{YELLOW!50}1 & \cellcolor{YELLOW!50}1 & \cellcolor{YELLOW!50}1 & \cellcolor{YELLOW!50}1 & \cellcolor{YELLOW!50}1 & \cellcolor{YELLOW!50}1 & \cellcolor{YELLOW!50}1 & \cellcolor{YELLOW!50}1 & \cellcolor{YELLOW!50}1 & \cellcolor{YELLOW!50}1 & \cellcolor{YELLOW!50}1 & \cellcolor{YELLOW!50}1 & \cellcolor{YELLOW!50}1 \\
\textbf{3: Speed limit 60} & \cellcolor{RED!50}2 & \cellcolor{RED!50}2 & \cellcolor{YELLOW!50}1 & - & \cellcolor{YELLOW!50}1 & \cellcolor{YELLOW!50}1 & \cellcolor{YELLOW!50}1 & \cellcolor{RED!50}2 & \cellcolor{RED!50}2 & \cellcolor{YELLOW!50}1 & \cellcolor{YELLOW!50}1 & \cellcolor{YELLOW!50}1 & \cellcolor{YELLOW!50}1 & \cellcolor{YELLOW!50}1 & \cellcolor{RED!50}2 & \cellcolor{YELLOW!50}1 & \cellcolor{YELLOW!50}1 & \cellcolor{RED!50}2 & \cellcolor{YELLOW!50}1 & \cellcolor{YELLOW!50}1 & \cellcolor{YELLOW!50}1 & \cellcolor{YELLOW!50}1 & \cellcolor{YELLOW!50}1 & \cellcolor{YELLOW!50}1 & \cellcolor{YELLOW!50}1 & \cellcolor{YELLOW!50}1 & \cellcolor{YELLOW!50}1 & \cellcolor{YELLOW!50}1 & \cellcolor{YELLOW!50}1 & \cellcolor{YELLOW!50}1 & \cellcolor{YELLOW!50}1 & \cellcolor{YELLOW!50}1 & \cellcolor{YELLOW!50}1 & \cellcolor{YELLOW!50}1 & \cellcolor{YELLOW!50}1 & \cellcolor{YELLOW!50}1 & \cellcolor{YELLOW!50}1 & \cellcolor{YELLOW!50}1 & \cellcolor{YELLOW!50}1 & \cellcolor{YELLOW!50}1 & \cellcolor{YELLOW!50}1 & \cellcolor{YELLOW!50}1 & \cellcolor{YELLOW!50}1 \\
\textbf{4: Speed limit 70} & \cellcolor{RED!50}2 & \cellcolor{RED!50}2 & \cellcolor{YELLOW!50}1 & \cellcolor{YELLOW!50}1 & - & \cellcolor{YELLOW!50}1 & \cellcolor{YELLOW!50}1 & \cellcolor{RED!50}2 & \cellcolor{RED!50}2 & \cellcolor{YELLOW!50}1 & \cellcolor{YELLOW!50}1 & \cellcolor{YELLOW!50}1 & \cellcolor{YELLOW!50}1 & \cellcolor{YELLOW!50}1 & \cellcolor{RED!50}2 & \cellcolor{YELLOW!50}1 & \cellcolor{YELLOW!50}1 & \cellcolor{RED!50}2 & \cellcolor{YELLOW!50}1 & \cellcolor{YELLOW!50}1 & \cellcolor{YELLOW!50}1 & \cellcolor{YELLOW!50}1 & \cellcolor{YELLOW!50}1 & \cellcolor{YELLOW!50}1 & \cellcolor{YELLOW!50}1 & \cellcolor{YELLOW!50}1 & \cellcolor{YELLOW!50}1 & \cellcolor{YELLOW!50}1 & \cellcolor{YELLOW!50}1 & \cellcolor{YELLOW!50}1 & \cellcolor{YELLOW!50}1 & \cellcolor{YELLOW!50}1 & \cellcolor{YELLOW!50}1 & \cellcolor{YELLOW!50}1 & \cellcolor{YELLOW!50}1 & \cellcolor{YELLOW!50}1 & \cellcolor{YELLOW!50}1 & \cellcolor{YELLOW!50}1 & \cellcolor{YELLOW!50}1 & \cellcolor{YELLOW!50}1 & \cellcolor{YELLOW!50}1 & \cellcolor{YELLOW!50}1 & \cellcolor{YELLOW!50}1 \\
\textbf{5: Speed limit 80} & \cellcolor{RED!50}2 & \cellcolor{RED!50}2 & \cellcolor{YELLOW!50}1 & \cellcolor{YELLOW!50}1 & \cellcolor{YELLOW!50}1 & - & \cellcolor{YELLOW!50}1 & \cellcolor{RED!50}2 & \cellcolor{RED!50}2 & \cellcolor{YELLOW!50}1 & \cellcolor{YELLOW!50}1 & \cellcolor{YELLOW!50}1 & \cellcolor{YELLOW!50}1 & \cellcolor{YELLOW!50}1 & \cellcolor{RED!50}2 & \cellcolor{YELLOW!50}1 & \cellcolor{YELLOW!50}1 & \cellcolor{RED!50}2 & \cellcolor{YELLOW!50}1 & \cellcolor{YELLOW!50}1 & \cellcolor{YELLOW!50}1 & \cellcolor{YELLOW!50}1 & \cellcolor{YELLOW!50}1 & \cellcolor{YELLOW!50}1 & \cellcolor{YELLOW!50}1 & \cellcolor{YELLOW!50}1 & \cellcolor{YELLOW!50}1 & \cellcolor{YELLOW!50}1 & \cellcolor{YELLOW!50}1 & \cellcolor{YELLOW!50}1 & \cellcolor{YELLOW!50}1 & \cellcolor{YELLOW!50}1 & \cellcolor{YELLOW!50}1 & \cellcolor{YELLOW!50}1 & \cellcolor{YELLOW!50}1 & \cellcolor{YELLOW!50}1 & \cellcolor{YELLOW!50}1 & \cellcolor{YELLOW!50}1 & \cellcolor{YELLOW!50}1 & \cellcolor{YELLOW!50}1 & \cellcolor{YELLOW!50}1 & \cellcolor{YELLOW!50}1 & \cellcolor{YELLOW!50}1 \\
\textbf{6: Restriction ends 80} & 0 & 0 & 0 & 0 & 0 & \cellcolor{RED!50}2 & - & \cellcolor{RED!50}2 & \cellcolor{RED!50}2 & \cellcolor{YELLOW!50}1 & \cellcolor{YELLOW!50}1 & \cellcolor{YELLOW!50}1 & \cellcolor{YELLOW!50}1 & \cellcolor{YELLOW!50}1 & \cellcolor{RED!50}2 & \cellcolor{YELLOW!50}1 & \cellcolor{YELLOW!50}1 & \cellcolor{YELLOW!50}1 & 0 & 0 & 0 & 0 & 0 & 0 & 0 & 0 & 0 & 0 & 0 & 0 & 0 & 0 & 0 & 0 & 0 & 0 & 0 & 0 & \cellcolor{YELLOW!50}1 & \cellcolor{YELLOW!50}1 & \cellcolor{YELLOW!50}1 & \cellcolor{YELLOW!50}1 & \cellcolor{YELLOW!50}1 \\
\textbf{7: Speed limit 100} & \cellcolor{RED!50}2 & \cellcolor{RED!50}2 & \cellcolor{RED!50}2 & \cellcolor{RED!50}2 & \cellcolor{RED!50}2 & \cellcolor{RED!50}2 & \cellcolor{YELLOW!50}1 & - & \cellcolor{YELLOW!50}1 & \cellcolor{YELLOW!50}1 & \cellcolor{YELLOW!50}1 & \cellcolor{YELLOW!50}1 & \cellcolor{YELLOW!50}1 & \cellcolor{YELLOW!50}1 & \cellcolor{RED!50}2 & \cellcolor{YELLOW!50}1 & \cellcolor{YELLOW!50}1 & \cellcolor{RED!50}2 & \cellcolor{YELLOW!50}1 & \cellcolor{YELLOW!50}1 & \cellcolor{YELLOW!50}1 & \cellcolor{YELLOW!50}1 & \cellcolor{YELLOW!50}1 & \cellcolor{YELLOW!50}1 & \cellcolor{YELLOW!50}1 & \cellcolor{YELLOW!50}1 & \cellcolor{YELLOW!50}1 & \cellcolor{YELLOW!50}1 & \cellcolor{YELLOW!50}1 & \cellcolor{YELLOW!50}1 & \cellcolor{YELLOW!50}1 & \cellcolor{YELLOW!50}1 & \cellcolor{YELLOW!50}1 & \cellcolor{YELLOW!50}1 & \cellcolor{YELLOW!50}1 & \cellcolor{YELLOW!50}1 & \cellcolor{YELLOW!50}1 & \cellcolor{YELLOW!50}1 & \cellcolor{YELLOW!50}1 & \cellcolor{YELLOW!50}1 & \cellcolor{YELLOW!50}1 & \cellcolor{YELLOW!50}1 & \cellcolor{YELLOW!50}1 \\
\textbf{8: Speed limit 120} & \cellcolor{RED!50}2 & \cellcolor{RED!50}2 & \cellcolor{RED!50}2 & \cellcolor{RED!50}2 & \cellcolor{RED!50}2 & \cellcolor{RED!50}2 & \cellcolor{YELLOW!50}1 & \cellcolor{YELLOW!50}1 & - & \cellcolor{YELLOW!50}1 & \cellcolor{YELLOW!50}1 & \cellcolor{YELLOW!50}1 & \cellcolor{YELLOW!50}1 & \cellcolor{YELLOW!50}1 & \cellcolor{RED!50}2 & \cellcolor{YELLOW!50}1 & \cellcolor{YELLOW!50}1 & \cellcolor{RED!50}2 & \cellcolor{YELLOW!50}1 & \cellcolor{YELLOW!50}1 & \cellcolor{YELLOW!50}1 & \cellcolor{YELLOW!50}1 & \cellcolor{YELLOW!50}1 & \cellcolor{YELLOW!50}1 & \cellcolor{YELLOW!50}1 & \cellcolor{YELLOW!50}1 & \cellcolor{YELLOW!50}1 & \cellcolor{YELLOW!50}1 & \cellcolor{YELLOW!50}1 & \cellcolor{YELLOW!50}1 & \cellcolor{YELLOW!50}1 & \cellcolor{YELLOW!50}1 & \cellcolor{YELLOW!50}1 & \cellcolor{YELLOW!50}1 & \cellcolor{YELLOW!50}1 & \cellcolor{YELLOW!50}1 & \cellcolor{YELLOW!50}1 & \cellcolor{YELLOW!50}1 & \cellcolor{YELLOW!50}1 & \cellcolor{YELLOW!50}1 & \cellcolor{YELLOW!50}1 & \cellcolor{YELLOW!50}1 & \cellcolor{YELLOW!50}1 \\
\textbf{9: No overtaking} & \cellcolor{YELLOW!50}1 & \cellcolor{YELLOW!50}1 & \cellcolor{YELLOW!50}1 & \cellcolor{YELLOW!50}1 & \cellcolor{YELLOW!50}1 & \cellcolor{YELLOW!50}1 & \cellcolor{YELLOW!50}1 & \cellcolor{YELLOW!50}1 & \cellcolor{YELLOW!50}1 & - & 0 & 0 & 0 & 0 & \cellcolor{YELLOW!50}1 & \cellcolor{YELLOW!50}1 & 0 & \cellcolor{YELLOW!50}1 & \cellcolor{YELLOW!50}1 & 0 & 0 & 0 & 0 & 0 & 0 & 0 & 0 & 0 & 0 & 0 & 0 & 0 & 0 & \cellcolor{YELLOW!50}1 & \cellcolor{YELLOW!50}1 & \cellcolor{YELLOW!50}1 & \cellcolor{YELLOW!50}1 & \cellcolor{YELLOW!50}1 & \cellcolor{YELLOW!50}1 & \cellcolor{YELLOW!50}1 & \cellcolor{YELLOW!50}1 & \cellcolor{YELLOW!50}1 & \cellcolor{YELLOW!50}1 \\
\textbf{10: No overtaking (trucks)} & \cellcolor{YELLOW!50}1 & \cellcolor{YELLOW!50}1 & \cellcolor{YELLOW!50}1 & \cellcolor{YELLOW!50}1 & \cellcolor{YELLOW!50}1 & \cellcolor{YELLOW!50}1 & \cellcolor{YELLOW!50}1 & \cellcolor{YELLOW!50}1 & \cellcolor{YELLOW!50}1 & 0 & - & 0 & 0 & 0 & \cellcolor{YELLOW!50}1 & \cellcolor{YELLOW!50}1 & 0 & \cellcolor{YELLOW!50}1 & \cellcolor{YELLOW!50}1 & 0 & 0 & 0 & 0 & 0 & 0 & 0 & 0 & 0 & 0 & 0 & 0 & 0 & 0 & \cellcolor{YELLOW!50}1 & \cellcolor{YELLOW!50}1 & \cellcolor{YELLOW!50}1 & \cellcolor{YELLOW!50}1 & \cellcolor{YELLOW!50}1 & \cellcolor{YELLOW!50}1 & \cellcolor{YELLOW!50}1 & \cellcolor{YELLOW!50}1 & \cellcolor{YELLOW!50}1 & \cellcolor{YELLOW!50}1 \\
\textbf{11: Priority next intersection} & \cellcolor{YELLOW!50}1 & \cellcolor{YELLOW!50}1 & \cellcolor{YELLOW!50}1 & \cellcolor{YELLOW!50}1 & \cellcolor{YELLOW!50}1 & \cellcolor{YELLOW!50}1 & \cellcolor{YELLOW!50}1 & \cellcolor{YELLOW!50}1 & \cellcolor{YELLOW!50}1 & 0 & 0 & - & 0 & 0 & \cellcolor{YELLOW!50}1 & \cellcolor{YELLOW!50}1 & 0 & \cellcolor{YELLOW!50}1 & \cellcolor{YELLOW!50}1 & 0 & 0 & 0 & 0 & 0 & 0 & 0 & 0 & 0 & 0 & 0 & 0 & 0 & 0 & \cellcolor{YELLOW!50}1 & \cellcolor{YELLOW!50}1 & \cellcolor{YELLOW!50}1 & \cellcolor{YELLOW!50}1 & \cellcolor{YELLOW!50}1 & \cellcolor{YELLOW!50}1 & \cellcolor{YELLOW!50}1 & \cellcolor{YELLOW!50}1 & \cellcolor{YELLOW!50}1 & \cellcolor{YELLOW!50}1 \\
\textbf{12: Priority road} & \cellcolor{YELLOW!50}1 & \cellcolor{YELLOW!50}1 & \cellcolor{YELLOW!50}1 & \cellcolor{YELLOW!50}1 & \cellcolor{YELLOW!50}1 & \cellcolor{YELLOW!50}1 & \cellcolor{YELLOW!50}1 & \cellcolor{YELLOW!50}1 & \cellcolor{YELLOW!50}1 & 0 & 0 & 0 & - & 0 & \cellcolor{YELLOW!50}1 & \cellcolor{YELLOW!50}1 & 0 & \cellcolor{YELLOW!50}1 & \cellcolor{YELLOW!50}1 & 0 & 0 & 0 & 0 & 0 & 0 & 0 & 0 & 0 & 0 & 0 & 0 & 0 & 0 & \cellcolor{YELLOW!50}1 & \cellcolor{YELLOW!50}1 & \cellcolor{YELLOW!50}1 & \cellcolor{YELLOW!50}1 & \cellcolor{YELLOW!50}1 & \cellcolor{YELLOW!50}1 & \cellcolor{YELLOW!50}1 & \cellcolor{YELLOW!50}1 & \cellcolor{YELLOW!50}1 & \cellcolor{YELLOW!50}1 \\
\textbf{13: Give way} & \cellcolor{YELLOW!50}1 & \cellcolor{YELLOW!50}1 & \cellcolor{YELLOW!50}1 & \cellcolor{YELLOW!50}1 & \cellcolor{YELLOW!50}1 & \cellcolor{YELLOW!50}1 & \cellcolor{YELLOW!50}1 & \cellcolor{YELLOW!50}1 & \cellcolor{YELLOW!50}1 & 0 & 0 & 0 & 0 & - & \cellcolor{YELLOW!50}1 & \cellcolor{YELLOW!50}1 & 0 & \cellcolor{YELLOW!50}1 & \cellcolor{YELLOW!50}1 & 0 & 0 & 0 & 0 & 0 & 0 & 0 & 0 & 0 & 0 & 0 & 0 & 0 & 0 & \cellcolor{YELLOW!50}1 & \cellcolor{YELLOW!50}1 & \cellcolor{YELLOW!50}1 & \cellcolor{YELLOW!50}1 & \cellcolor{YELLOW!50}1 & \cellcolor{YELLOW!50}1 & \cellcolor{YELLOW!50}1 & \cellcolor{YELLOW!50}1 & \cellcolor{YELLOW!50}1 & \cellcolor{YELLOW!50}1 \\
\textbf{14: Stop} & \cellcolor{RED!50}2 & \cellcolor{RED!50}2 & \cellcolor{RED!50}2 & \cellcolor{RED!50}2 & \cellcolor{RED!50}2 & \cellcolor{RED!50}2 & \cellcolor{RED!50}2 & \cellcolor{RED!50}2 & \cellcolor{RED!50}2 & \cellcolor{YELLOW!50}1 & \cellcolor{YELLOW!50}1 & \cellcolor{YELLOW!50}1 & \cellcolor{YELLOW!50}1 & \cellcolor{YELLOW!50}1 & - & \cellcolor{YELLOW!50}1 & \cellcolor{YELLOW!50}1 & 0 & 0 & \cellcolor{YELLOW!50}1 & \cellcolor{YELLOW!50}1 & \cellcolor{YELLOW!50}1 & \cellcolor{YELLOW!50}1 & \cellcolor{YELLOW!50}1 & \cellcolor{YELLOW!50}1 & \cellcolor{YELLOW!50}1 & \cellcolor{YELLOW!50}1 & 0 & 0 & 0 & \cellcolor{YELLOW!50}1 & \cellcolor{YELLOW!50}1 & \cellcolor{YELLOW!50}1 & 0 & 0 & 0 & 0 & 0 & \cellcolor{YELLOW!50}1 & \cellcolor{YELLOW!50}1 & \cellcolor{YELLOW!50}1 & \cellcolor{YELLOW!50}1 & \cellcolor{YELLOW!50}1 \\
\textbf{15: No traffic both ways} & \cellcolor{YELLOW!50}1 & \cellcolor{YELLOW!50}1 & \cellcolor{YELLOW!50}1 & \cellcolor{YELLOW!50}1 & \cellcolor{YELLOW!50}1 & \cellcolor{YELLOW!50}1 & \cellcolor{YELLOW!50}1 & \cellcolor{YELLOW!50}1 & \cellcolor{YELLOW!50}1 & \cellcolor{YELLOW!50}1 & \cellcolor{YELLOW!50}1 & \cellcolor{YELLOW!50}1 & \cellcolor{YELLOW!50}1 & \cellcolor{YELLOW!50}1 & \cellcolor{YELLOW!50}1 & - & \cellcolor{YELLOW!50}1 & \cellcolor{YELLOW!50}1 & \cellcolor{YELLOW!50}1 & \cellcolor{YELLOW!50}1 & \cellcolor{YELLOW!50}1 & \cellcolor{YELLOW!50}1 & \cellcolor{YELLOW!50}1 & \cellcolor{YELLOW!50}1 & \cellcolor{YELLOW!50}1 & \cellcolor{YELLOW!50}1 & \cellcolor{YELLOW!50}1 & \cellcolor{YELLOW!50}1 & \cellcolor{YELLOW!50}1 & \cellcolor{YELLOW!50}1 & \cellcolor{YELLOW!50}1 & \cellcolor{YELLOW!50}1 & \cellcolor{YELLOW!50}1 & \cellcolor{YELLOW!50}1 & \cellcolor{YELLOW!50}1 & \cellcolor{YELLOW!50}1 & \cellcolor{YELLOW!50}1 & \cellcolor{YELLOW!50}1 & \cellcolor{YELLOW!50}1 & \cellcolor{YELLOW!50}1 & \cellcolor{YELLOW!50}1 & \cellcolor{YELLOW!50}1 & \cellcolor{YELLOW!50}1 \\
\textbf{16: No trucks} & \cellcolor{YELLOW!50}1 & \cellcolor{YELLOW!50}1 & \cellcolor{YELLOW!50}1 & \cellcolor{YELLOW!50}1 & \cellcolor{YELLOW!50}1 & \cellcolor{YELLOW!50}1 & \cellcolor{YELLOW!50}1 & \cellcolor{YELLOW!50}1 & \cellcolor{YELLOW!50}1 & 0 & 0 & 0 & 0 & 0 & \cellcolor{YELLOW!50}1 & \cellcolor{YELLOW!50}1 & - & \cellcolor{YELLOW!50}1 & \cellcolor{YELLOW!50}1 & 0 & 0 & 0 & 0 & 0 & 0 & 0 & 0 & 0 & 0 & 0 & 0 & 0 & 0 & \cellcolor{YELLOW!50}1 & \cellcolor{YELLOW!50}1 & \cellcolor{YELLOW!50}1 & \cellcolor{YELLOW!50}1 & \cellcolor{YELLOW!50}1 & \cellcolor{YELLOW!50}1 & \cellcolor{YELLOW!50}1 & \cellcolor{YELLOW!50}1 & \cellcolor{YELLOW!50}1 & \cellcolor{YELLOW!50}1 \\
\textbf{17: No entry} & \cellcolor{RED!50}2 & \cellcolor{RED!50}2 & \cellcolor{RED!50}2 & \cellcolor{RED!50}2 & \cellcolor{RED!50}2 & \cellcolor{RED!50}2 & \cellcolor{RED!50}2 & \cellcolor{RED!50}2 & \cellcolor{RED!50}2 & \cellcolor{YELLOW!50}1 & \cellcolor{YELLOW!50}1 & \cellcolor{YELLOW!50}1 & \cellcolor{YELLOW!50}1 & \cellcolor{YELLOW!50}1 & 0 & \cellcolor{YELLOW!50}1 & \cellcolor{YELLOW!50}1 & - & \cellcolor{YELLOW!50}1 & \cellcolor{YELLOW!50}1 & \cellcolor{YELLOW!50}1 & \cellcolor{YELLOW!50}1 & \cellcolor{YELLOW!50}1 & \cellcolor{YELLOW!50}1 & \cellcolor{YELLOW!50}1 & \cellcolor{YELLOW!50}1 & \cellcolor{YELLOW!50}1 & \cellcolor{YELLOW!50}1 & \cellcolor{YELLOW!50}1 & \cellcolor{YELLOW!50}1 & \cellcolor{YELLOW!50}1 & \cellcolor{YELLOW!50}1 & \cellcolor{YELLOW!50}1 & \cellcolor{RED!50}2 & \cellcolor{RED!50}2 & \cellcolor{RED!50}2 & \cellcolor{RED!50}2 & \cellcolor{RED!50}2 & \cellcolor{YELLOW!50}1 & \cellcolor{YELLOW!50}1 & \cellcolor{YELLOW!50}1 & \cellcolor{YELLOW!50}1 & \cellcolor{YELLOW!50}1 \\
\textbf{18: Danger} & \cellcolor{YELLOW!50}1 & \cellcolor{YELLOW!50}1 & \cellcolor{YELLOW!50}1 & \cellcolor{YELLOW!50}1 & \cellcolor{YELLOW!50}1 & \cellcolor{YELLOW!50}1 & 0 & \cellcolor{YELLOW!50}1 & \cellcolor{YELLOW!50}1 & 0 & 0 & 0 & 0 & 0 & 0 & \cellcolor{YELLOW!50}1 & 0 & 0 & - & 0 & 0 & 0 & 0 & 0 & 0 & 0 & 0 & 0 & 0 & 0 & 0 & 0 & 0 & 0 & 0 & 0 & 0 & 0 & \cellcolor{YELLOW!50}1 & \cellcolor{YELLOW!50}1 & \cellcolor{YELLOW!50}1 & \cellcolor{YELLOW!50}1 & \cellcolor{YELLOW!50}1 \\
\textbf{19: Bend left} & \cellcolor{YELLOW!50}1 & \cellcolor{YELLOW!50}1 & \cellcolor{YELLOW!50}1 & \cellcolor{YELLOW!50}1 & \cellcolor{YELLOW!50}1 & \cellcolor{YELLOW!50}1 & 0 & \cellcolor{YELLOW!50}1 & \cellcolor{YELLOW!50}1 & 0 & 0 & 0 & 0 & 0 & \cellcolor{YELLOW!50}1 & \cellcolor{YELLOW!50}1 & 0 & \cellcolor{YELLOW!50}1 & 0 & - & 0 & 0 & 0 & 0 & 0 & 0 & 0 & 0 & 0 & 0 & 0 & 0 & 0 & \cellcolor{YELLOW!50}1 & \cellcolor{YELLOW!50}1 & \cellcolor{YELLOW!50}1 & \cellcolor{YELLOW!50}1 & \cellcolor{YELLOW!50}1 & \cellcolor{YELLOW!50}1 & \cellcolor{YELLOW!50}1 & \cellcolor{YELLOW!50}1 & \cellcolor{YELLOW!50}1 & \cellcolor{YELLOW!50}1 \\
\textbf{20: Bend right} & \cellcolor{YELLOW!50}1 & \cellcolor{YELLOW!50}1 & \cellcolor{YELLOW!50}1 & \cellcolor{YELLOW!50}1 & \cellcolor{YELLOW!50}1 & \cellcolor{YELLOW!50}1 & 0 & \cellcolor{YELLOW!50}1 & \cellcolor{YELLOW!50}1 & 0 & 0 & 0 & 0 & 0 & \cellcolor{YELLOW!50}1 & \cellcolor{YELLOW!50}1 & 0 & \cellcolor{YELLOW!50}1 & 0 & 0 & - & 0 & 0 & 0 & 0 & 0 & 0 & 0 & 0 & 0 & 0 & 0 & 0 & \cellcolor{YELLOW!50}1 & \cellcolor{YELLOW!50}1 & \cellcolor{YELLOW!50}1 & \cellcolor{YELLOW!50}1 & \cellcolor{YELLOW!50}1 & \cellcolor{YELLOW!50}1 & \cellcolor{YELLOW!50}1 & \cellcolor{YELLOW!50}1 & \cellcolor{YELLOW!50}1 & \cellcolor{YELLOW!50}1 \\
\textbf{21: Bend} & \cellcolor{YELLOW!50}1 & \cellcolor{YELLOW!50}1 & \cellcolor{YELLOW!50}1 & \cellcolor{YELLOW!50}1 & \cellcolor{YELLOW!50}1 & \cellcolor{YELLOW!50}1 & 0 & \cellcolor{YELLOW!50}1 & \cellcolor{YELLOW!50}1 & 0 & 0 & 0 & 0 & 0 & \cellcolor{YELLOW!50}1 & \cellcolor{YELLOW!50}1 & 0 & \cellcolor{YELLOW!50}1 & 0 & 0 & 0 & - & 0 & 0 & 0 & 0 & 0 & 0 & 0 & 0 & 0 & 0 & 0 & \cellcolor{YELLOW!50}1 & \cellcolor{YELLOW!50}1 & \cellcolor{YELLOW!50}1 & \cellcolor{YELLOW!50}1 & \cellcolor{YELLOW!50}1 & \cellcolor{YELLOW!50}1 & \cellcolor{YELLOW!50}1 & \cellcolor{YELLOW!50}1 & \cellcolor{YELLOW!50}1 & \cellcolor{YELLOW!50}1 \\
\textbf{22: Uneven road} & \cellcolor{YELLOW!50}1 & \cellcolor{YELLOW!50}1 & \cellcolor{YELLOW!50}1 & \cellcolor{YELLOW!50}1 & \cellcolor{YELLOW!50}1 & \cellcolor{YELLOW!50}1 & 0 & \cellcolor{YELLOW!50}1 & \cellcolor{YELLOW!50}1 & 0 & 0 & 0 & 0 & 0 & \cellcolor{YELLOW!50}1 & \cellcolor{YELLOW!50}1 & 0 & \cellcolor{YELLOW!50}1 & 0 & 0 & 0 & 0 & - & 0 & 0 & 0 & 0 & 0 & 0 & 0 & 0 & 0 & 0 & \cellcolor{YELLOW!50}1 & \cellcolor{YELLOW!50}1 & \cellcolor{YELLOW!50}1 & \cellcolor{YELLOW!50}1 & \cellcolor{YELLOW!50}1 & \cellcolor{YELLOW!50}1 & \cellcolor{YELLOW!50}1 & \cellcolor{YELLOW!50}1 & \cellcolor{YELLOW!50}1 & \cellcolor{YELLOW!50}1 \\
\textbf{23: Slippery road} & \cellcolor{YELLOW!50}1 & \cellcolor{YELLOW!50}1 & \cellcolor{YELLOW!50}1 & \cellcolor{YELLOW!50}1 & \cellcolor{YELLOW!50}1 & \cellcolor{YELLOW!50}1 & 0 & \cellcolor{YELLOW!50}1 & \cellcolor{YELLOW!50}1 & 0 & 0 & 0 & 0 & 0 & \cellcolor{YELLOW!50}1 & \cellcolor{YELLOW!50}1 & 0 & \cellcolor{YELLOW!50}1 & 0 & 0 & 0 & 0 & 0 & - & 0 & 0 & 0 & 0 & 0 & 0 & 0 & 0 & 0 & \cellcolor{YELLOW!50}1 & \cellcolor{YELLOW!50}1 & \cellcolor{YELLOW!50}1 & \cellcolor{YELLOW!50}1 & \cellcolor{YELLOW!50}1 & \cellcolor{YELLOW!50}1 & \cellcolor{YELLOW!50}1 & \cellcolor{YELLOW!50}1 & \cellcolor{YELLOW!50}1 & \cellcolor{YELLOW!50}1 \\
\textbf{24: Road narrows} & \cellcolor{YELLOW!50}1 & \cellcolor{YELLOW!50}1 & \cellcolor{YELLOW!50}1 & \cellcolor{YELLOW!50}1 & \cellcolor{YELLOW!50}1 & \cellcolor{YELLOW!50}1 & 0 & \cellcolor{YELLOW!50}1 & \cellcolor{YELLOW!50}1 & 0 & 0 & 0 & 0 & 0 & \cellcolor{YELLOW!50}1 & \cellcolor{YELLOW!50}1 & 0 & \cellcolor{YELLOW!50}1 & 0 & 0 & 0 & 0 & 0 & 0 & - & 0 & 0 & 0 & 0 & 0 & 0 & 0 & 0 & \cellcolor{YELLOW!50}1 & \cellcolor{YELLOW!50}1 & \cellcolor{YELLOW!50}1 & \cellcolor{YELLOW!50}1 & \cellcolor{YELLOW!50}1 & \cellcolor{YELLOW!50}1 & \cellcolor{YELLOW!50}1 & \cellcolor{YELLOW!50}1 & \cellcolor{YELLOW!50}1 & \cellcolor{YELLOW!50}1 \\
\textbf{25: Construction} & \cellcolor{YELLOW!50}1 & \cellcolor{YELLOW!50}1 & \cellcolor{YELLOW!50}1 & \cellcolor{YELLOW!50}1 & \cellcolor{YELLOW!50}1 & \cellcolor{YELLOW!50}1 & 0 & \cellcolor{YELLOW!50}1 & \cellcolor{YELLOW!50}1 & 0 & 0 & 0 & 0 & 0 & \cellcolor{YELLOW!50}1 & \cellcolor{YELLOW!50}1 & 0 & \cellcolor{YELLOW!50}1 & 0 & 0 & 0 & 0 & 0 & 0 & 0 & - & 0 & 0 & 0 & 0 & 0 & 0 & 0 & \cellcolor{YELLOW!50}1 & \cellcolor{YELLOW!50}1 & \cellcolor{YELLOW!50}1 & \cellcolor{YELLOW!50}1 & \cellcolor{YELLOW!50}1 & \cellcolor{YELLOW!50}1 & \cellcolor{YELLOW!50}1 & \cellcolor{YELLOW!50}1 & \cellcolor{YELLOW!50}1 & \cellcolor{YELLOW!50}1 \\
\textbf{26: Traffic signal} & \cellcolor{YELLOW!50}1 & \cellcolor{YELLOW!50}1 & \cellcolor{YELLOW!50}1 & \cellcolor{YELLOW!50}1 & \cellcolor{YELLOW!50}1 & \cellcolor{YELLOW!50}1 & 0 & \cellcolor{YELLOW!50}1 & \cellcolor{YELLOW!50}1 & 0 & 0 & 0 & 0 & 0 & \cellcolor{YELLOW!50}1 & \cellcolor{YELLOW!50}1 & 0 & \cellcolor{YELLOW!50}1 & 0 & 0 & 0 & 0 & 0 & 0 & 0 & 0 & - & 0 & 0 & 0 & 0 & 0 & 0 & \cellcolor{YELLOW!50}1 & \cellcolor{YELLOW!50}1 & \cellcolor{YELLOW!50}1 & \cellcolor{YELLOW!50}1 & \cellcolor{YELLOW!50}1 & \cellcolor{YELLOW!50}1 & \cellcolor{YELLOW!50}1 & \cellcolor{YELLOW!50}1 & \cellcolor{YELLOW!50}1 & \cellcolor{YELLOW!50}1 \\
\textbf{27: Pedestrian crossing} & \cellcolor{YELLOW!50}1 & \cellcolor{YELLOW!50}1 & \cellcolor{YELLOW!50}1 & \cellcolor{YELLOW!50}1 & \cellcolor{YELLOW!50}1 & \cellcolor{YELLOW!50}1 & 0 & \cellcolor{YELLOW!50}1 & \cellcolor{YELLOW!50}1 & 0 & 0 & 0 & 0 & 0 & 0 & \cellcolor{YELLOW!50}1 & 0 & 0 & 0 & 0 & 0 & 0 & 0 & 0 & 0 & 0 & 0 & - & 0 & 0 & 0 & 0 & 0 & 0 & 0 & 0 & 0 & 0 & \cellcolor{YELLOW!50}1 & \cellcolor{YELLOW!50}1 & \cellcolor{YELLOW!50}1 & \cellcolor{YELLOW!50}1 & \cellcolor{YELLOW!50}1 \\
\textbf{28: School crossing} & \cellcolor{YELLOW!50}1 & \cellcolor{YELLOW!50}1 & \cellcolor{YELLOW!50}1 & \cellcolor{YELLOW!50}1 & \cellcolor{YELLOW!50}1 & \cellcolor{YELLOW!50}1 & 0 & \cellcolor{YELLOW!50}1 & \cellcolor{YELLOW!50}1 & 0 & 0 & 0 & 0 & 0 & 0 & \cellcolor{YELLOW!50}1 & 0 & 0 & 0 & 0 & 0 & 0 & 0 & 0 & 0 & 0 & 0 & 0 & - & 0 & 0 & 0 & 0 & 0 & 0 & 0 & 0 & 0 & \cellcolor{YELLOW!50}1 & \cellcolor{YELLOW!50}1 & \cellcolor{YELLOW!50}1 & \cellcolor{YELLOW!50}1 & \cellcolor{YELLOW!50}1 \\
\textbf{29: Cycles crossing} & \cellcolor{YELLOW!50}1 & \cellcolor{YELLOW!50}1 & \cellcolor{YELLOW!50}1 & \cellcolor{YELLOW!50}1 & \cellcolor{YELLOW!50}1 & \cellcolor{YELLOW!50}1 & 0 & \cellcolor{YELLOW!50}1 & \cellcolor{YELLOW!50}1 & 0 & 0 & 0 & 0 & 0 & 0 & \cellcolor{YELLOW!50}1 & 0 & 0 & 0 & 0 & 0 & 0 & 0 & 0 & 0 & 0 & 0 & 0 & 0 & - & 0 & 0 & 0 & 0 & 0 & 0 & 0 & 0 & \cellcolor{YELLOW!50}1 & \cellcolor{YELLOW!50}1 & \cellcolor{YELLOW!50}1 & \cellcolor{YELLOW!50}1 & \cellcolor{YELLOW!50}1 \\
\textbf{30: Snow} & \cellcolor{YELLOW!50}1 & \cellcolor{YELLOW!50}1 & \cellcolor{YELLOW!50}1 & \cellcolor{YELLOW!50}1 & \cellcolor{YELLOW!50}1 & \cellcolor{YELLOW!50}1 & \cellcolor{YELLOW!50}1 & \cellcolor{YELLOW!50}1 & \cellcolor{YELLOW!50}1 & 0 & 0 & 0 & 0 & 0 & \cellcolor{YELLOW!50}1 & \cellcolor{YELLOW!50}1 & 0 & \cellcolor{YELLOW!50}1 & 0 & 0 & 0 & 0 & 0 & 0 & 0 & 0 & 0 & 0 & 0 & 0 & - & 0 & 0 & \cellcolor{YELLOW!50}1 & \cellcolor{YELLOW!50}1 & \cellcolor{YELLOW!50}1 & \cellcolor{YELLOW!50}1 & \cellcolor{YELLOW!50}1 & \cellcolor{YELLOW!50}1 & \cellcolor{YELLOW!50}1 & \cellcolor{YELLOW!50}1 & \cellcolor{YELLOW!50}1 & \cellcolor{YELLOW!50}1 \\
\textbf{31: Animals} & \cellcolor{YELLOW!50}1 & \cellcolor{YELLOW!50}1 & \cellcolor{YELLOW!50}1 & \cellcolor{YELLOW!50}1 & \cellcolor{YELLOW!50}1 & \cellcolor{YELLOW!50}1 & \cellcolor{YELLOW!50}1 & \cellcolor{YELLOW!50}1 & \cellcolor{YELLOW!50}1 & 0 & 0 & 0 & 0 & 0 & \cellcolor{YELLOW!50}1 & \cellcolor{YELLOW!50}1 & 0 & \cellcolor{YELLOW!50}1 & 0 & 0 & 0 & 0 & 0 & 0 & 0 & 0 & 0 & 0 & 0 & 0 & 0 & - & 0 & \cellcolor{YELLOW!50}1 & \cellcolor{YELLOW!50}1 & \cellcolor{YELLOW!50}1 & \cellcolor{YELLOW!50}1 & \cellcolor{YELLOW!50}1 & \cellcolor{YELLOW!50}1 & \cellcolor{YELLOW!50}1 & \cellcolor{YELLOW!50}1 & \cellcolor{YELLOW!50}1 & \cellcolor{YELLOW!50}1 \\
\textbf{32: Restriction ends} & \cellcolor{YELLOW!50}1 & \cellcolor{YELLOW!50}1 & \cellcolor{YELLOW!50}1 & \cellcolor{YELLOW!50}1 & \cellcolor{YELLOW!50}1 & \cellcolor{YELLOW!50}1 & \cellcolor{YELLOW!50}1 & \cellcolor{YELLOW!50}1 & \cellcolor{YELLOW!50}1 & 0 & 0 & 0 & 0 & 0 & \cellcolor{YELLOW!50}1 & \cellcolor{YELLOW!50}1 & 0 & \cellcolor{YELLOW!50}1 & 0 & 0 & 0 & 0 & 0 & 0 & 0 & 0 & 0 & 0 & 0 & 0 & 0 & 0 & - & \cellcolor{YELLOW!50}1 & \cellcolor{YELLOW!50}1 & \cellcolor{YELLOW!50}1 & \cellcolor{YELLOW!50}1 & \cellcolor{YELLOW!50}1 & \cellcolor{YELLOW!50}1 & \cellcolor{YELLOW!50}1 & \cellcolor{YELLOW!50}1 & \cellcolor{YELLOW!50}1 & \cellcolor{YELLOW!50}1 \\
\textbf{33: Go right} & 0 & 0 & 0 & 0 & 0 & 0 & 0 & 0 & 0 & \cellcolor{YELLOW!50}1 & \cellcolor{YELLOW!50}1 & \cellcolor{YELLOW!50}1 & \cellcolor{YELLOW!50}1 & \cellcolor{YELLOW!50}1 & 0 & \cellcolor{YELLOW!50}1 & \cellcolor{YELLOW!50}1 & 0 & 0 & \cellcolor{YELLOW!50}1 & \cellcolor{YELLOW!50}1 & \cellcolor{YELLOW!50}1 & \cellcolor{YELLOW!50}1 & \cellcolor{YELLOW!50}1 & \cellcolor{YELLOW!50}1 & \cellcolor{YELLOW!50}1 & \cellcolor{YELLOW!50}1 & 0 & 0 & 0 & \cellcolor{YELLOW!50}1 & \cellcolor{YELLOW!50}1 & \cellcolor{YELLOW!50}1 & - & \cellcolor{RED!50}2 & \cellcolor{RED!50}2 & \cellcolor{RED!50}2 & \cellcolor{RED!50}2 & \cellcolor{YELLOW!50}1 & \cellcolor{YELLOW!50}1 & \cellcolor{YELLOW!50}1 & \cellcolor{YELLOW!50}1 & \cellcolor{YELLOW!50}1 \\
\textbf{34: Go left} & 0 & 0 & 0 & 0 & 0 & 0 & 0 & 0 & 0 & \cellcolor{YELLOW!50}1 & \cellcolor{YELLOW!50}1 & \cellcolor{YELLOW!50}1 & \cellcolor{YELLOW!50}1 & \cellcolor{YELLOW!50}1 & 0 & \cellcolor{YELLOW!50}1 & \cellcolor{YELLOW!50}1 & 0 & 0 & \cellcolor{YELLOW!50}1 & \cellcolor{YELLOW!50}1 & \cellcolor{YELLOW!50}1 & \cellcolor{YELLOW!50}1 & \cellcolor{YELLOW!50}1 & \cellcolor{YELLOW!50}1 & \cellcolor{YELLOW!50}1 & \cellcolor{YELLOW!50}1 & 0 & 0 & 0 & \cellcolor{YELLOW!50}1 & \cellcolor{YELLOW!50}1 & \cellcolor{YELLOW!50}1 & \cellcolor{RED!50}2 & - & \cellcolor{RED!50}2 & \cellcolor{RED!50}2 & \cellcolor{RED!50}2 & \cellcolor{YELLOW!50}1 & \cellcolor{YELLOW!50}1 & \cellcolor{YELLOW!50}1 & \cellcolor{YELLOW!50}1 & \cellcolor{YELLOW!50}1 \\
\textbf{35: Go straight} & 0 & 0 & 0 & 0 & 0 & 0 & 0 & 0 & 0 & \cellcolor{YELLOW!50}1 & \cellcolor{YELLOW!50}1 & \cellcolor{YELLOW!50}1 & \cellcolor{YELLOW!50}1 & \cellcolor{YELLOW!50}1 & 0 & \cellcolor{YELLOW!50}1 & \cellcolor{YELLOW!50}1 & 0 & 0 & \cellcolor{YELLOW!50}1 & \cellcolor{YELLOW!50}1 & \cellcolor{YELLOW!50}1 & \cellcolor{YELLOW!50}1 & \cellcolor{YELLOW!50}1 & \cellcolor{YELLOW!50}1 & \cellcolor{YELLOW!50}1 & \cellcolor{YELLOW!50}1 & 0 & 0 & 0 & \cellcolor{YELLOW!50}1 & \cellcolor{YELLOW!50}1 & \cellcolor{YELLOW!50}1 & \cellcolor{RED!50}2 & \cellcolor{RED!50}2 & - & \cellcolor{RED!50}2 & \cellcolor{RED!50}2 & \cellcolor{YELLOW!50}1 & \cellcolor{YELLOW!50}1 & \cellcolor{YELLOW!50}1 & \cellcolor{YELLOW!50}1 & \cellcolor{YELLOW!50}1 \\
\textbf{36: Go right or straight} & 0 & 0 & 0 & 0 & 0 & 0 & 0 & 0 & 0 & \cellcolor{YELLOW!50}1 & \cellcolor{YELLOW!50}1 & \cellcolor{YELLOW!50}1 & \cellcolor{YELLOW!50}1 & \cellcolor{YELLOW!50}1 & 0 & \cellcolor{YELLOW!50}1 & \cellcolor{YELLOW!50}1 & 0 & 0 & \cellcolor{YELLOW!50}1 & \cellcolor{YELLOW!50}1 & \cellcolor{YELLOW!50}1 & \cellcolor{YELLOW!50}1 & \cellcolor{YELLOW!50}1 & \cellcolor{YELLOW!50}1 & \cellcolor{YELLOW!50}1 & \cellcolor{YELLOW!50}1 & 0 & 0 & 0 & \cellcolor{YELLOW!50}1 & \cellcolor{YELLOW!50}1 & \cellcolor{YELLOW!50}1 & \cellcolor{RED!50}2 & \cellcolor{RED!50}2 & \cellcolor{RED!50}2 & - & \cellcolor{RED!50}2 & \cellcolor{YELLOW!50}1 & \cellcolor{YELLOW!50}1 & \cellcolor{YELLOW!50}1 & \cellcolor{YELLOW!50}1 & \cellcolor{YELLOW!50}1 \\
\textbf{37: Go left or straight} & 0 & 0 & 0 & 0 & 0 & 0 & 0 & 0 & 0 & \cellcolor{YELLOW!50}1 & \cellcolor{YELLOW!50}1 & \cellcolor{YELLOW!50}1 & \cellcolor{YELLOW!50}1 & \cellcolor{YELLOW!50}1 & 0 & \cellcolor{YELLOW!50}1 & \cellcolor{YELLOW!50}1 & 0 & 0 & \cellcolor{YELLOW!50}1 & \cellcolor{YELLOW!50}1 & \cellcolor{YELLOW!50}1 & \cellcolor{YELLOW!50}1 & \cellcolor{YELLOW!50}1 & \cellcolor{YELLOW!50}1 & \cellcolor{YELLOW!50}1 & \cellcolor{YELLOW!50}1 & 0 & 0 & 0 & \cellcolor{YELLOW!50}1 & \cellcolor{YELLOW!50}1 & \cellcolor{YELLOW!50}1 & \cellcolor{RED!50}2 & \cellcolor{RED!50}2 & \cellcolor{RED!50}2 & \cellcolor{RED!50}2 & - & \cellcolor{YELLOW!50}1 & \cellcolor{YELLOW!50}1 & \cellcolor{YELLOW!50}1 & \cellcolor{YELLOW!50}1 & \cellcolor{YELLOW!50}1 \\
\textbf{38: Keep right} & \cellcolor{YELLOW!50}1 & \cellcolor{YELLOW!50}1 & \cellcolor{YELLOW!50}1 & \cellcolor{YELLOW!50}1 & \cellcolor{YELLOW!50}1 & \cellcolor{YELLOW!50}1 & \cellcolor{YELLOW!50}1 & \cellcolor{YELLOW!50}1 & \cellcolor{YELLOW!50}1 & \cellcolor{YELLOW!50}1 & \cellcolor{YELLOW!50}1 & \cellcolor{YELLOW!50}1 & \cellcolor{YELLOW!50}1 & \cellcolor{YELLOW!50}1 & \cellcolor{YELLOW!50}1 & \cellcolor{YELLOW!50}1 & \cellcolor{YELLOW!50}1 & \cellcolor{YELLOW!50}1 & \cellcolor{YELLOW!50}1 & \cellcolor{YELLOW!50}1 & \cellcolor{YELLOW!50}1 & \cellcolor{YELLOW!50}1 & \cellcolor{YELLOW!50}1 & \cellcolor{YELLOW!50}1 & \cellcolor{YELLOW!50}1 & \cellcolor{YELLOW!50}1 & \cellcolor{YELLOW!50}1 & \cellcolor{YELLOW!50}1 & \cellcolor{YELLOW!50}1 & \cellcolor{YELLOW!50}1 & \cellcolor{YELLOW!50}1 & \cellcolor{YELLOW!50}1 & \cellcolor{YELLOW!50}1 & \cellcolor{YELLOW!50}1 & \cellcolor{YELLOW!50}1 & \cellcolor{YELLOW!50}1 & \cellcolor{YELLOW!50}1 & \cellcolor{YELLOW!50}1 & - & \cellcolor{YELLOW!50}1 & \cellcolor{YELLOW!50}1 & \cellcolor{YELLOW!50}1 & \cellcolor{YELLOW!50}1 \\
\textbf{39: Keep left} & \cellcolor{YELLOW!50}1 & \cellcolor{YELLOW!50}1 & \cellcolor{YELLOW!50}1 & \cellcolor{YELLOW!50}1 & \cellcolor{YELLOW!50}1 & \cellcolor{YELLOW!50}1 & \cellcolor{YELLOW!50}1 & \cellcolor{YELLOW!50}1 & \cellcolor{YELLOW!50}1 & \cellcolor{YELLOW!50}1 & \cellcolor{YELLOW!50}1 & \cellcolor{YELLOW!50}1 & \cellcolor{YELLOW!50}1 & \cellcolor{YELLOW!50}1 & \cellcolor{YELLOW!50}1 & \cellcolor{YELLOW!50}1 & \cellcolor{YELLOW!50}1 & \cellcolor{YELLOW!50}1 & \cellcolor{YELLOW!50}1 & \cellcolor{YELLOW!50}1 & \cellcolor{YELLOW!50}1 & \cellcolor{YELLOW!50}1 & \cellcolor{YELLOW!50}1 & \cellcolor{YELLOW!50}1 & \cellcolor{YELLOW!50}1 & \cellcolor{YELLOW!50}1 & \cellcolor{YELLOW!50}1 & \cellcolor{YELLOW!50}1 & \cellcolor{YELLOW!50}1 & \cellcolor{YELLOW!50}1 & \cellcolor{YELLOW!50}1 & \cellcolor{YELLOW!50}1 & \cellcolor{YELLOW!50}1 & \cellcolor{YELLOW!50}1 & \cellcolor{YELLOW!50}1 & \cellcolor{YELLOW!50}1 & \cellcolor{YELLOW!50}1 & \cellcolor{YELLOW!50}1 & \cellcolor{YELLOW!50}1 & - & \cellcolor{YELLOW!50}1 & \cellcolor{YELLOW!50}1 & \cellcolor{YELLOW!50}1 \\
\textbf{40: Roundabout} & \cellcolor{YELLOW!50}1 & \cellcolor{YELLOW!50}1 & \cellcolor{YELLOW!50}1 & \cellcolor{YELLOW!50}1 & \cellcolor{YELLOW!50}1 & \cellcolor{YELLOW!50}1 & \cellcolor{YELLOW!50}1 & \cellcolor{YELLOW!50}1 & \cellcolor{YELLOW!50}1 & \cellcolor{YELLOW!50}1 & \cellcolor{YELLOW!50}1 & \cellcolor{YELLOW!50}1 & \cellcolor{YELLOW!50}1 & \cellcolor{YELLOW!50}1 & \cellcolor{YELLOW!50}1 & \cellcolor{YELLOW!50}1 & \cellcolor{YELLOW!50}1 & \cellcolor{YELLOW!50}1 & \cellcolor{YELLOW!50}1 & \cellcolor{YELLOW!50}1 & \cellcolor{YELLOW!50}1 & \cellcolor{YELLOW!50}1 & \cellcolor{YELLOW!50}1 & \cellcolor{YELLOW!50}1 & \cellcolor{YELLOW!50}1 & \cellcolor{YELLOW!50}1 & \cellcolor{YELLOW!50}1 & \cellcolor{YELLOW!50}1 & \cellcolor{YELLOW!50}1 & \cellcolor{YELLOW!50}1 & \cellcolor{YELLOW!50}1 & \cellcolor{YELLOW!50}1 & \cellcolor{YELLOW!50}1 & \cellcolor{YELLOW!50}1 & \cellcolor{YELLOW!50}1 & \cellcolor{YELLOW!50}1 & \cellcolor{YELLOW!50}1 & \cellcolor{YELLOW!50}1 & \cellcolor{YELLOW!50}1 & \cellcolor{YELLOW!50}1 & - & \cellcolor{YELLOW!50}1 & \cellcolor{YELLOW!50}1 \\
\textbf{41: Restriction ends} & \cellcolor{YELLOW!50}1 & \cellcolor{YELLOW!50}1 & \cellcolor{YELLOW!50}1 & \cellcolor{YELLOW!50}1 & \cellcolor{YELLOW!50}1 & \cellcolor{YELLOW!50}1 & \cellcolor{YELLOW!50}1 & \cellcolor{YELLOW!50}1 & \cellcolor{YELLOW!50}1 & \cellcolor{YELLOW!50}1 & \cellcolor{YELLOW!50}1 & \cellcolor{YELLOW!50}1 & \cellcolor{YELLOW!50}1 & \cellcolor{YELLOW!50}1 & \cellcolor{YELLOW!50}1 & \cellcolor{YELLOW!50}1 & \cellcolor{YELLOW!50}1 & \cellcolor{YELLOW!50}1 & \cellcolor{YELLOW!50}1 & \cellcolor{YELLOW!50}1 & \cellcolor{YELLOW!50}1 & \cellcolor{YELLOW!50}1 & \cellcolor{YELLOW!50}1 & \cellcolor{YELLOW!50}1 & \cellcolor{YELLOW!50}1 & \cellcolor{YELLOW!50}1 & \cellcolor{YELLOW!50}1 & \cellcolor{YELLOW!50}1 & \cellcolor{YELLOW!50}1 & \cellcolor{YELLOW!50}1 & \cellcolor{YELLOW!50}1 & \cellcolor{YELLOW!50}1 & \cellcolor{YELLOW!50}1 & \cellcolor{YELLOW!50}1 & \cellcolor{YELLOW!50}1 & \cellcolor{YELLOW!50}1 & \cellcolor{YELLOW!50}1 & \cellcolor{YELLOW!50}1 & \cellcolor{YELLOW!50}1 & \cellcolor{YELLOW!50}1 & \cellcolor{YELLOW!50}1 & - & \cellcolor{YELLOW!50}1 \\
\textbf{42: Restriction ends (trucks)} & \cellcolor{YELLOW!50}1 & \cellcolor{YELLOW!50}1 & \cellcolor{YELLOW!50}1 & \cellcolor{YELLOW!50}1 & \cellcolor{YELLOW!50}1 & \cellcolor{YELLOW!50}1 & \cellcolor{YELLOW!50}1 & \cellcolor{YELLOW!50}1 & \cellcolor{YELLOW!50}1 & \cellcolor{YELLOW!50}1 & \cellcolor{YELLOW!50}1 & \cellcolor{YELLOW!50}1 & \cellcolor{YELLOW!50}1 & \cellcolor{YELLOW!50}1 & \cellcolor{YELLOW!50}1 & \cellcolor{YELLOW!50}1 & \cellcolor{YELLOW!50}1 & \cellcolor{YELLOW!50}1 & \cellcolor{YELLOW!50}1 & \cellcolor{YELLOW!50}1 & \cellcolor{YELLOW!50}1 & \cellcolor{YELLOW!50}1 & \cellcolor{YELLOW!50}1 & \cellcolor{YELLOW!50}1 & \cellcolor{YELLOW!50}1 & \cellcolor{YELLOW!50}1 & \cellcolor{YELLOW!50}1 & \cellcolor{YELLOW!50}1 & \cellcolor{YELLOW!50}1 & \cellcolor{YELLOW!50}1 & \cellcolor{YELLOW!50}1 & \cellcolor{YELLOW!50}1 & \cellcolor{YELLOW!50}1 & \cellcolor{YELLOW!50}1 & \cellcolor{YELLOW!50}1 & \cellcolor{YELLOW!50}1 & \cellcolor{YELLOW!50}1 & \cellcolor{YELLOW!50}1 & \cellcolor{YELLOW!50}1 & \cellcolor{YELLOW!50}1 & \cellcolor{YELLOW!50}1 & \cellcolor{YELLOW!50}1 & - \\
\end{tabular*}
\end{sidewaystable}

\begin{sidewaystable}
\setlength{\tabcolsep}{5pt}
\caption{Severity level definitions for misclassifications of traffic signs in ArTS}\label{SecondSeverityMatrix}
\begin{tabular*}{\textwidth}{@{\extracolsep\fill}l*{24}{c}}
\toprule
\multirow{2}{22mm}{\textbf{Ground Truth}} & \multicolumn{24}{c}{\textbf{Prediction}} \\
\cmidrule{2-25}
 & \textbf{1} & \textbf{2} & \textbf{3} & \textbf{4} & \textbf{5} & \textbf{6} & \textbf{7} & \textbf{8} & \textbf{9} & \textbf{10} & \textbf{11} & \textbf{12} & \textbf{13} & \textbf{14} & \textbf{15} & \textbf{16} & \textbf{17} & \textbf{18} & \textbf{19} & \textbf{20} & \textbf{21} & \textbf{22} & \textbf{23} & \textbf{24} \\
\midrule
1: Front or Right   & 0  & \cellcolor{RED!50}2  & \cellcolor{YELLOW!50}1  & \cellcolor{YELLOW!50}1  & \cellcolor{YELLOW!50}1  & \cellcolor{YELLOW!50}1  & \cellcolor{YELLOW!50}1  & \cellcolor{YELLOW!50}1  & \cellcolor{YELLOW!50}1  & \cellcolor{YELLOW!50}1  & \cellcolor{YELLOW!50}1  & \cellcolor{YELLOW!50}1  & \cellcolor{YELLOW!50}1  & \cellcolor{YELLOW!50}1  & \cellcolor{YELLOW!50}1  & \cellcolor{YELLOW!50}1  & \cellcolor{YELLOW!50}1  & \cellcolor{YELLOW!50}1  & \cellcolor{YELLOW!50}1  & \cellcolor{YELLOW!50}1  & \cellcolor{YELLOW!50}1  & \cellcolor{YELLOW!50}1  & \cellcolor{YELLOW!50}1  & \cellcolor{YELLOW!50}1 \\
2: Front or Left    & \cellcolor{RED!50}2  & 0  & \cellcolor{YELLOW!50}1  & \cellcolor{YELLOW!50}1  & \cellcolor{YELLOW!50}1  & \cellcolor{YELLOW!50}1  & \cellcolor{YELLOW!50}1  & \cellcolor{YELLOW!50}1  & \cellcolor{YELLOW!50}1  & \cellcolor{YELLOW!50}1  & \cellcolor{YELLOW!50}1  & \cellcolor{YELLOW!50}1  & \cellcolor{YELLOW!50}1  & \cellcolor{YELLOW!50}1  & \cellcolor{YELLOW!50}1  & \cellcolor{YELLOW!50}1  & \cellcolor{YELLOW!50}1  & \cellcolor{YELLOW!50}1  & \cellcolor{YELLOW!50}1  & \cellcolor{YELLOW!50}1  & \cellcolor{YELLOW!50}1  & \cellcolor{YELLOW!50}1  & \cellcolor{YELLOW!50}1  & \cellcolor{YELLOW!50}1 \\
3: Road Hump        & \cellcolor{YELLOW!50}1  & \cellcolor{YELLOW!50}1  & 0  & 0  & 0  & \cellcolor{YELLOW!50}1  & 0  & 0  & 0  & 0  & 0  & 0  & 0  & \cellcolor{YELLOW!50}1  & 0  & 0  & 0  & 0  & \cellcolor{YELLOW!50}1  & \cellcolor{YELLOW!50}1  & \cellcolor{YELLOW!50}1  & \cellcolor{YELLOW!50}1  & \cellcolor{YELLOW!50}1  & \cellcolor{YELLOW!50}1 \\
4: Left Turn        & \cellcolor{YELLOW!50}1  & \cellcolor{YELLOW!50}1  & 0  & 0  & 0  & \cellcolor{YELLOW!50}1  & 0  & 0  & 0  & 0  & 0  & 0  & 0  & \cellcolor{YELLOW!50}1  & 0  & 0  & 0  & 0  & \cellcolor{YELLOW!50}1  & \cellcolor{YELLOW!50}1  & \cellcolor{YELLOW!50}1  & \cellcolor{YELLOW!50}1  & \cellcolor{YELLOW!50}1  & \cellcolor{YELLOW!50}1 \\
5: Narrow From Right & \cellcolor{YELLOW!50}1  & \cellcolor{YELLOW!50}1  & 0  & 0  & 0  & \cellcolor{YELLOW!50}1  & 0  & 0  & 0  & 0  & 0  & 0  & 0  & \cellcolor{YELLOW!50}1  & 0  & 0  & 0  & 0  & \cellcolor{YELLOW!50}1  & \cellcolor{YELLOW!50}1  & \cellcolor{YELLOW!50}1  & \cellcolor{YELLOW!50}1  & \cellcolor{YELLOW!50}1  & \cellcolor{YELLOW!50}1 \\
6: No U Turn        & \cellcolor{YELLOW!50}1  & \cellcolor{YELLOW!50}1  & \cellcolor{YELLOW!50}1  & \cellcolor{YELLOW!50}1  & \cellcolor{YELLOW!50}1  & 0  & \cellcolor{YELLOW!50}1  & \cellcolor{YELLOW!50}1  & \cellcolor{YELLOW!50}1  & \cellcolor{YELLOW!50}1  & \cellcolor{YELLOW!50}1  & \cellcolor{YELLOW!50}1  & \cellcolor{YELLOW!50}1  & \cellcolor{YELLOW!50}1  & \cellcolor{YELLOW!50}1  & \cellcolor{YELLOW!50}1  & \cellcolor{YELLOW!50}1  & \cellcolor{YELLOW!50}1  & \cellcolor{YELLOW!50}1  & \cellcolor{YELLOW!50}1  & \cellcolor{YELLOW!50}1  & \cellcolor{YELLOW!50}1  & \cellcolor{YELLOW!50}1  & \cellcolor{YELLOW!50}1 \\
7: Parking          & \cellcolor{YELLOW!50}1  & \cellcolor{YELLOW!50}1  & 0  & 0  & 0  & \cellcolor{YELLOW!50}1  & 0  & 0  & 0  & 0  & 0  & 0  & 0  & \cellcolor{YELLOW!50}1  & 0  & 0  & 0  & 0  & \cellcolor{YELLOW!50}1  & \cellcolor{YELLOW!50}1  & \cellcolor{YELLOW!50}1  & \cellcolor{YELLOW!50}1  & \cellcolor{YELLOW!50}1  & \cellcolor{YELLOW!50}1 \\
8: No Hom           & \cellcolor{YELLOW!50}1  & \cellcolor{YELLOW!50}1  & 0  & 0  & 0  & \cellcolor{YELLOW!50}1  & 0  & 0  & 0  & 0  & 0  & 0  & 0  & \cellcolor{YELLOW!50}1  & 0  & 0  & 0  & 0  & \cellcolor{YELLOW!50}1  & \cellcolor{YELLOW!50}1  & \cellcolor{YELLOW!50}1  & \cellcolor{YELLOW!50}1  & \cellcolor{YELLOW!50}1  & \cellcolor{YELLOW!50}1 \\
9: Right Tum         & \cellcolor{YELLOW!50}1  & \cellcolor{YELLOW!50}1  & 0  & 0  & 0  & \cellcolor{YELLOW!50}1  & 0  & 0  & 0  & 0  & 0  & 0  & 0  & \cellcolor{YELLOW!50}1  & 0  & 0  & 0  & 0  & \cellcolor{YELLOW!50}1  & \cellcolor{YELLOW!50}1  & \cellcolor{YELLOW!50}1  & \cellcolor{YELLOW!50}1  & \cellcolor{YELLOW!50}1  & \cellcolor{YELLOW!50}1 \\
10: Narrow From Left & \cellcolor{YELLOW!50}1  & \cellcolor{YELLOW!50}1  & 0  & 0  & 0  & \cellcolor{YELLOW!50}1  & 0  & 0  & 0  & 0  & 0  & 0  & 0  & \cellcolor{YELLOW!50}1  & 0  & 0  & 0  & 0  & \cellcolor{YELLOW!50}1  & \cellcolor{YELLOW!50}1  & \cellcolor{YELLOW!50}1  & \cellcolor{YELLOW!50}1  & \cellcolor{YELLOW!50}1  & \cellcolor{YELLOW!50}1 \\
11: U Tum             & \cellcolor{YELLOW!50}1  & \cellcolor{YELLOW!50}1  & 0  & 0  & 0  & \cellcolor{YELLOW!50}1  & 0  & 0  & 0  & 0  & 0  & 0  & 0  & \cellcolor{YELLOW!50}1  & 0  & 0  & 0  & 0  & \cellcolor{YELLOW!50}1  & \cellcolor{YELLOW!50}1  & \cellcolor{YELLOW!50}1  & \cellcolor{YELLOW!50}1  & \cellcolor{YELLOW!50}1  & \cellcolor{YELLOW!50}1 \\
12: Rotor             & \cellcolor{YELLOW!50}1  & \cellcolor{YELLOW!50}1  & 0  & 0  & 0  & \cellcolor{YELLOW!50}1  & 0  & 0  & 0  & 0  & 0  & 0  & 0  & \cellcolor{YELLOW!50}1  & 0  & 0  & 0  & 0  & \cellcolor{YELLOW!50}1  & \cellcolor{YELLOW!50}1  & \cellcolor{YELLOW!50}1  & \cellcolor{YELLOW!50}1  & \cellcolor{YELLOW!50}1  & \cellcolor{YELLOW!50}1 \\
13: Slow              & \cellcolor{YELLOW!50}1  & \cellcolor{YELLOW!50}1  & 0  & 0  & 0  & \cellcolor{YELLOW!50}1  & 0  & 0  & 0  & 0  & 0  & 0  & 0  & \cellcolor{YELLOW!50}1  & 0  & 0  & 0  & 0  & \cellcolor{YELLOW!50}1  & \cellcolor{YELLOW!50}1  & \cellcolor{YELLOW!50}1  & \cellcolor{YELLOW!50}1  & \cellcolor{YELLOW!50}1  & \cellcolor{YELLOW!50}1 \\
14: Stop              & \cellcolor{YELLOW!50}1  & \cellcolor{YELLOW!50}1  & \cellcolor{YELLOW!50}1  & \cellcolor{YELLOW!50}1  & \cellcolor{YELLOW!50}1  & \cellcolor{YELLOW!50}1  & \cellcolor{YELLOW!50}1  & \cellcolor{YELLOW!50}1  & \cellcolor{YELLOW!50}1  & \cellcolor{YELLOW!50}1  & \cellcolor{YELLOW!50}1  & \cellcolor{YELLOW!50}1  & \cellcolor{YELLOW!50}1  & 0  & \cellcolor{YELLOW!50}1  & \cellcolor{YELLOW!50}1  & \cellcolor{YELLOW!50}1  & \cellcolor{YELLOW!50}1  & \cellcolor{RED!50}2  & \cellcolor{RED!50}2  & \cellcolor{RED!50}2  & \cellcolor{RED!50}2  & \cellcolor{RED!50}2  & \cellcolor{RED!50}2 \\
15: No Overtaking     & \cellcolor{YELLOW!50}1  & \cellcolor{YELLOW!50}1  & 0  & 0  & 0  & \cellcolor{YELLOW!50}1  & 0  & 0  & 0  & 0  & 0  & 0  & 0  & \cellcolor{YELLOW!50}1  & 0  & 0  & 0  & 0  & \cellcolor{YELLOW!50}1  & \cellcolor{YELLOW!50}1  & \cellcolor{YELLOW!50}1  & \cellcolor{YELLOW!50}1  & \cellcolor{YELLOW!50}1  & \cellcolor{YELLOW!50}1 \\
16: Right or Left     & \cellcolor{YELLOW!50}1  & \cellcolor{YELLOW!50}1  & 0  & 0  & 0  & \cellcolor{YELLOW!50}1  & 0  & 0  & 0  & 0  & 0  & 0  & 0  & \cellcolor{YELLOW!50}1  & 0  & 0  & 0  & 0  & \cellcolor{YELLOW!50}1  & \cellcolor{YELLOW!50}1  & \cellcolor{YELLOW!50}1  & \cellcolor{YELLOW!50}1  & \cellcolor{YELLOW!50}1  & \cellcolor{YELLOW!50}1 \\
17: Pedestrian crossing & \cellcolor{YELLOW!50}1  & \cellcolor{YELLOW!50}1  & 0  & 0  & 0  & \cellcolor{YELLOW!50}1  & 0  & 0  & 0  & 0  & 0  & 0  & 0  & \cellcolor{YELLOW!50}1  & 0  & 0  & 0  & 0  & \cellcolor{YELLOW!50}1  & \cellcolor{YELLOW!50}1  & \cellcolor{YELLOW!50}1  & \cellcolor{YELLOW!50}1  & \cellcolor{YELLOW!50}1  & \cellcolor{YELLOW!50}1 \\
18: No Parking        & \cellcolor{YELLOW!50}1  & \cellcolor{YELLOW!50}1  & 0  & 0  & 0  & \cellcolor{YELLOW!50}1  & 0  & 0  & 0  & 0  & 0  & 0  & 0  & \cellcolor{YELLOW!50}1  & 0  & 0  & 0  & 0  & \cellcolor{YELLOW!50}1  & \cellcolor{YELLOW!50}1  & \cellcolor{YELLOW!50}1  & \cellcolor{YELLOW!50}1  & \cellcolor{YELLOW!50}1  & \cellcolor{YELLOW!50}1 \\
19: Speed 30          & \cellcolor{YELLOW!50}1  & \cellcolor{YELLOW!50}1  & \cellcolor{YELLOW!50}1  & \cellcolor{YELLOW!50}1  & \cellcolor{YELLOW!50}1  & \cellcolor{YELLOW!50}1  & \cellcolor{YELLOW!50}1  & \cellcolor{YELLOW!50}1  & \cellcolor{YELLOW!50}1  & \cellcolor{YELLOW!50}1  & \cellcolor{YELLOW!50}1  & \cellcolor{YELLOW!50}1  & \cellcolor{YELLOW!50}1  & \cellcolor{RED!50}2  & \cellcolor{YELLOW!50}1  & \cellcolor{YELLOW!50}1  & \cellcolor{YELLOW!50}1  & \cellcolor{YELLOW!50}1  & 0  & \cellcolor{YELLOW!50}1  & \cellcolor{RED!50}2  & \cellcolor{RED!50}2  & \cellcolor{RED!50}2  & \cellcolor{RED!50}2 \\
20: Speed 40          & \cellcolor{YELLOW!50}1  & \cellcolor{YELLOW!50}1  & \cellcolor{YELLOW!50}1  & \cellcolor{YELLOW!50}1  & \cellcolor{YELLOW!50}1  & \cellcolor{YELLOW!50}1  & \cellcolor{YELLOW!50}1  & \cellcolor{YELLOW!50}1  & \cellcolor{YELLOW!50}1  & \cellcolor{YELLOW!50}1  & \cellcolor{YELLOW!50}1  & \cellcolor{YELLOW!50}1  & \cellcolor{YELLOW!50}1  & \cellcolor{RED!50}2  & \cellcolor{YELLOW!50}1  & \cellcolor{YELLOW!50}1  & \cellcolor{YELLOW!50}1  & \cellcolor{YELLOW!50}1  & \cellcolor{YELLOW!50}1  & 0  & \cellcolor{YELLOW!50}1  & \cellcolor{RED!50}2  & \cellcolor{RED!50}2  & \cellcolor{RED!50}2 \\
21: Speed 50          & \cellcolor{YELLOW!50}1  & \cellcolor{YELLOW!50}1  & \cellcolor{YELLOW!50}1  & \cellcolor{YELLOW!50}1  & \cellcolor{YELLOW!50}1  & \cellcolor{YELLOW!50}1  & \cellcolor{YELLOW!50}1  & \cellcolor{YELLOW!50}1  & \cellcolor{YELLOW!50}1  & \cellcolor{YELLOW!50}1  & \cellcolor{YELLOW!50}1  & \cellcolor{YELLOW!50}1  & \cellcolor{YELLOW!50}1  & \cellcolor{RED!50}2  & \cellcolor{YELLOW!50}1  & \cellcolor{YELLOW!50}1  & \cellcolor{YELLOW!50}1  & \cellcolor{YELLOW!50}1  & \cellcolor{RED!50}2  & \cellcolor{YELLOW!50}1  & 0  & \cellcolor{YELLOW!50}1  & \cellcolor{RED!50}2  & \cellcolor{RED!50}2 \\
22: Speed 60          & \cellcolor{YELLOW!50}1  & \cellcolor{YELLOW!50}1  & \cellcolor{YELLOW!50}1  & \cellcolor{YELLOW!50}1  & \cellcolor{YELLOW!50}1  & \cellcolor{YELLOW!50}1  & \cellcolor{YELLOW!50}1  & \cellcolor{YELLOW!50}1  & \cellcolor{YELLOW!50}1  & \cellcolor{YELLOW!50}1  & \cellcolor{YELLOW!50}1  & \cellcolor{YELLOW!50}1  & \cellcolor{YELLOW!50}1  & \cellcolor{RED!50}2  & \cellcolor{YELLOW!50}1  & \cellcolor{YELLOW!50}1  & \cellcolor{YELLOW!50}1  & \cellcolor{YELLOW!50}1  & \cellcolor{RED!50}2  & \cellcolor{RED!50}2  & \cellcolor{YELLOW!50}1  & 0  & \cellcolor{YELLOW!50}1  & \cellcolor{RED!50}2 \\
23: Speed 80          & \cellcolor{YELLOW!50}1  & \cellcolor{YELLOW!50}1  & \cellcolor{YELLOW!50}1  & \cellcolor{YELLOW!50}1  & \cellcolor{YELLOW!50}1  & \cellcolor{YELLOW!50}1  & \cellcolor{YELLOW!50}1  & \cellcolor{YELLOW!50}1  & \cellcolor{YELLOW!50}1  & \cellcolor{YELLOW!50}1  & \cellcolor{YELLOW!50}1  & \cellcolor{YELLOW!50}1  & \cellcolor{YELLOW!50}1  & \cellcolor{RED!50}2  & \cellcolor{YELLOW!50}1  & \cellcolor{YELLOW!50}1  & \cellcolor{YELLOW!50}1  & \cellcolor{YELLOW!50}1  & \cellcolor{RED!50}2  & \cellcolor{RED!50}2  & \cellcolor{RED!50}2  & \cellcolor{YELLOW!50}1  & 0  & \cellcolor{YELLOW!50}1 \\
24: Speed 100         & \cellcolor{YELLOW!50}1  & \cellcolor{YELLOW!50}1  & \cellcolor{YELLOW!50}1  & \cellcolor{YELLOW!50}1  & \cellcolor{YELLOW!50}1  & \cellcolor{YELLOW!50}1  & \cellcolor{YELLOW!50}1  & \cellcolor{YELLOW!50}1  & \cellcolor{YELLOW!50}1  & \cellcolor{YELLOW!50}1  & \cellcolor{YELLOW!50}1  & \cellcolor{YELLOW!50}1  & \cellcolor{YELLOW!50}1  & \cellcolor{RED!50}2  & \cellcolor{YELLOW!50}1  & \cellcolor{YELLOW!50}1  & \cellcolor{YELLOW!50}1  & \cellcolor{YELLOW!50}1  & \cellcolor{RED!50}2  & \cellcolor{RED!50}2  & \cellcolor{RED!50}2  & \cellcolor{RED!50}2  & \cellcolor{YELLOW!50}1  & 0 \\
\bottomrule
\end{tabular*}
\end{sidewaystable}

\end{appendices}



\end{document}